\documentclass[10pt,twocolumn,letterpaper]{article}
\usepackage{wacv}
\usepackage{times}
\usepackage{epsfig}
\usepackage{graphicx}
\usepackage{amsmath}
\usepackage{amssymb}
\usepackage{mathtools}
\usepackage{subfigure}
\usepackage{color}
\usepackage{soul}
\usepackage{algorithm}
\usepackage{algpseudocode}
\usepackage{enumitem}
\usepackage{url}
\usepackage{bm}
\usepackage[T1]{fontenc} 
\usepackage{xspace}
\usepackage{caption}
\usepackage{gensymb}
\usepackage{epstopdf}

\newcommand{\qed}{\nobreak \ifvmode \relax \else
      \ifdim\lastskip<1.5em \hskip-\lastskip
      \hskip1.5em plus0em minus0.5em \fi \nobreak
      \vrule height0.75em width0.5em depth0.25em\fi}

\newcommand{\cA}{\mathcal{A}}

\newcommand{\bp}{\mathbf{p}}
\newcommand{\bu}{\mathbf{u}}
\newcommand{\bv}{\mathbf{v}}

\newcommand{\bs}{\mathbf{s}}

\newcommand{\br}{\mathbf{r}}

     \usepackage{color}
     \usepackage{xcolor}

\newcommand{\inv}{^{\raisebox{.2ex}{$\scriptscriptstyle-\!1$}}}

\newcommand{\bT}{\mathbf{T}}
\newcommand{\be}{\mathbf{e}}
\newcommand{\scA}{_{\raisebox{.1ex}{$\scriptscriptstyle\cA$}}}
\def\NoNumber#1{{\def\alglinenumber##1{}\State #1}\addtocounter{ALG@line}{-1}}

\usepackage[pagebackref=true,breaklinks=true,letterpaper=true,colorlinks,bookmarks=false]{hyperref}

\wacvfinalcopy 

\def\wacvPaperID{364} 

\ifwacvfinal\fi
\begin{document}

\title{Minimal Solvers for Monocular Rolling Shutter Compensation under Ackermann Motion}

\author{Pulak Purkait, Christopher Zach\\
Toshiba Research Europe, 
Cambridge, UK\\
{\tt\small pulak.isi@gmail.com}
}

\maketitle
\ifwacvfinal\thispagestyle{empty}\fi

\maketitle
\thispagestyle{empty}

\begin{abstract}
Modern automotive vehicles are often equipped with a budget commercial rolling shutter camera. These devices often produce distorted images due to the inter-row delay of the camera while capturing the image. Recent methods for monocular rolling shutter motion compensation utilize blur kernel and the straightness property of line segments. However, these methods are limited to handling rotational motion and also are not fast enough to operate in real time. In this paper, we propose a minimal solver for the rolling shutter motion compensation which assumes known vertical direction of the camera. Thanks to the Ackermann motion model of vehicles which consists of only two motion parameters, and two parameters for the simplified depth assumption that lead to a 4-line algorithm. The proposed minimal solver estimates the rolling shutter camera motion efficiently and accurately. The extensive experiments on real and simulated datasets demonstrate the benefits of our approach in terms of qualitative and quantitative results. 
\end{abstract}

\section{Introduction}
Recently, automotive driver-less vehicles have sparked a lot of vision research and unlocked the demands for the real time solutions for a number of unsolved problems in computer vision. While a commercial budget camera is an attractive choice of these vehicles, a significant  distortion could be observed in the captured images. These cameras are generally built upon CMOS sensors, which possess a prevalent mechanism widely known as \emph{rolling shutter} (RS). In contrast to \emph{global shutter} (GS) camera, it captures the scene in a row-wise fashion from top to bottom with a constant inter-row delay. The RS imaging acquires apparent camera motion for different rows and violates the  properties of the perspective camera model. This causes noticeable and prominent distortions. 

\begin{figure}[!ht]
\begin{center} 
\begin{tabular}{c@{\hspace{0em}}c}
\includegraphics[width=0.23\textwidth]{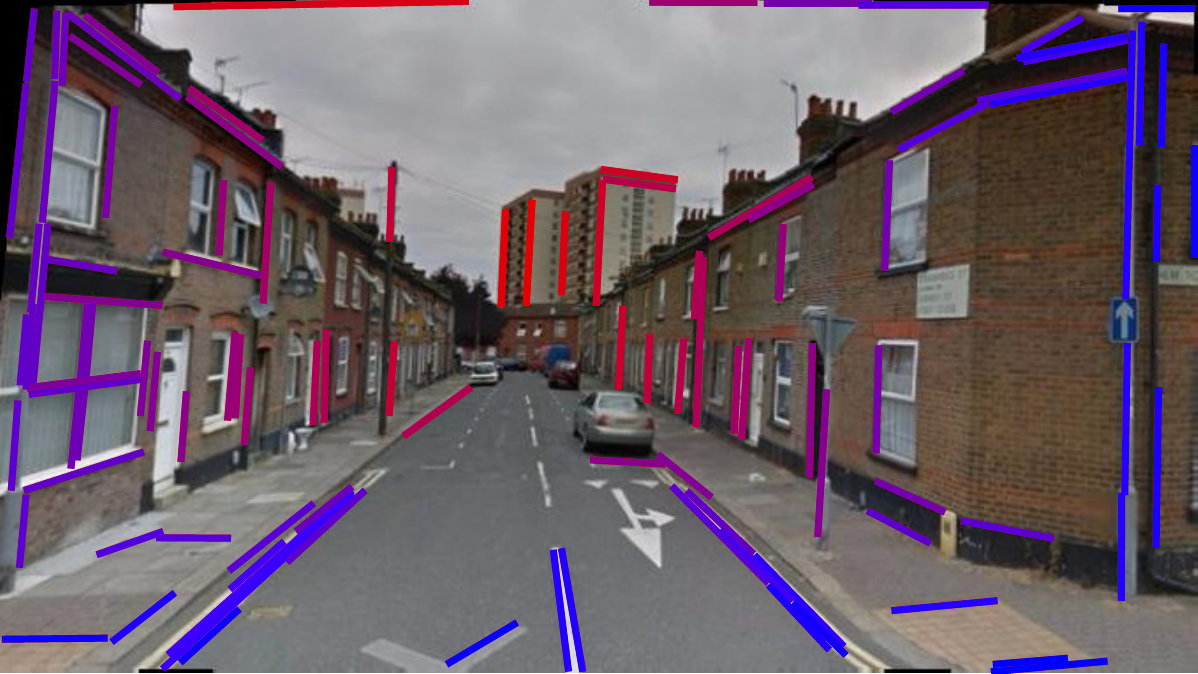} &
\includegraphics[width=0.23\textwidth]{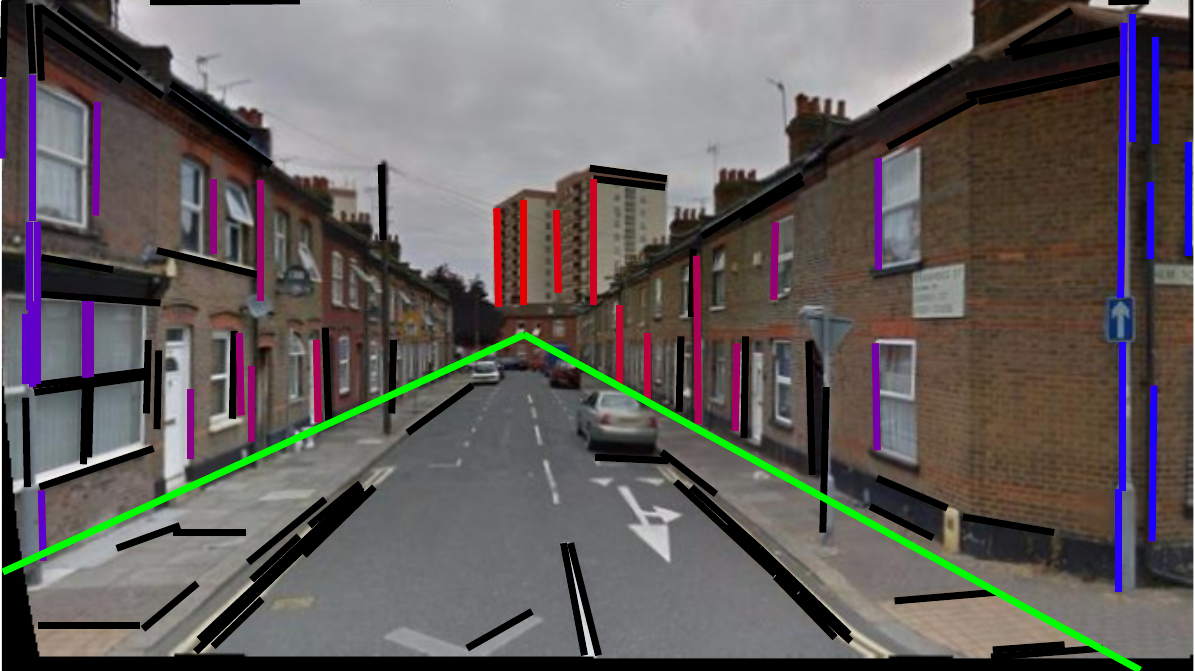} \\ 
\footnotesize (a) A synthetic rolling shutter image & \footnotesize (b) Motion compensated image \\ 
\includegraphics[width=0.23\textwidth]{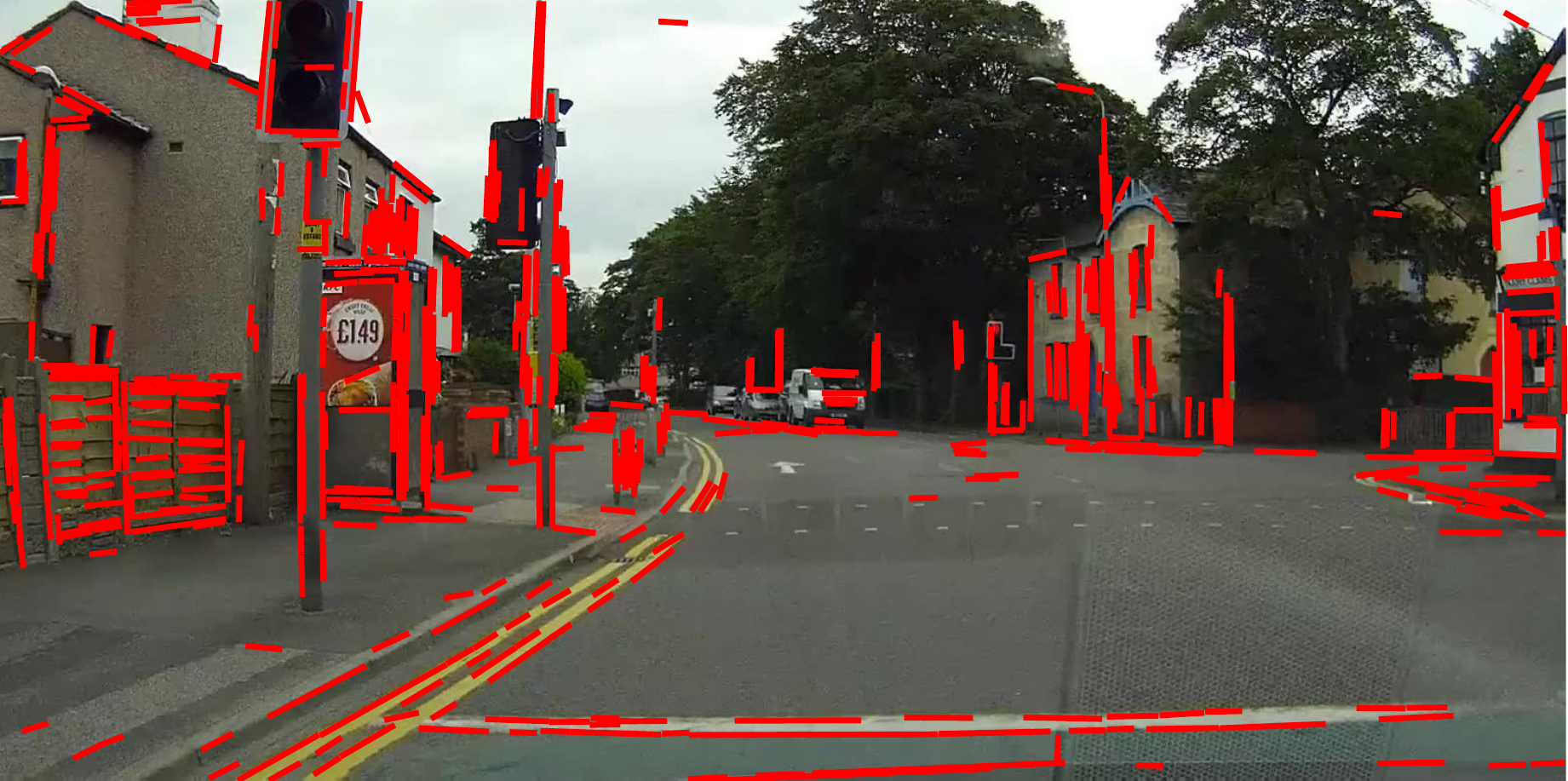} &
\includegraphics[width=0.23\textwidth]{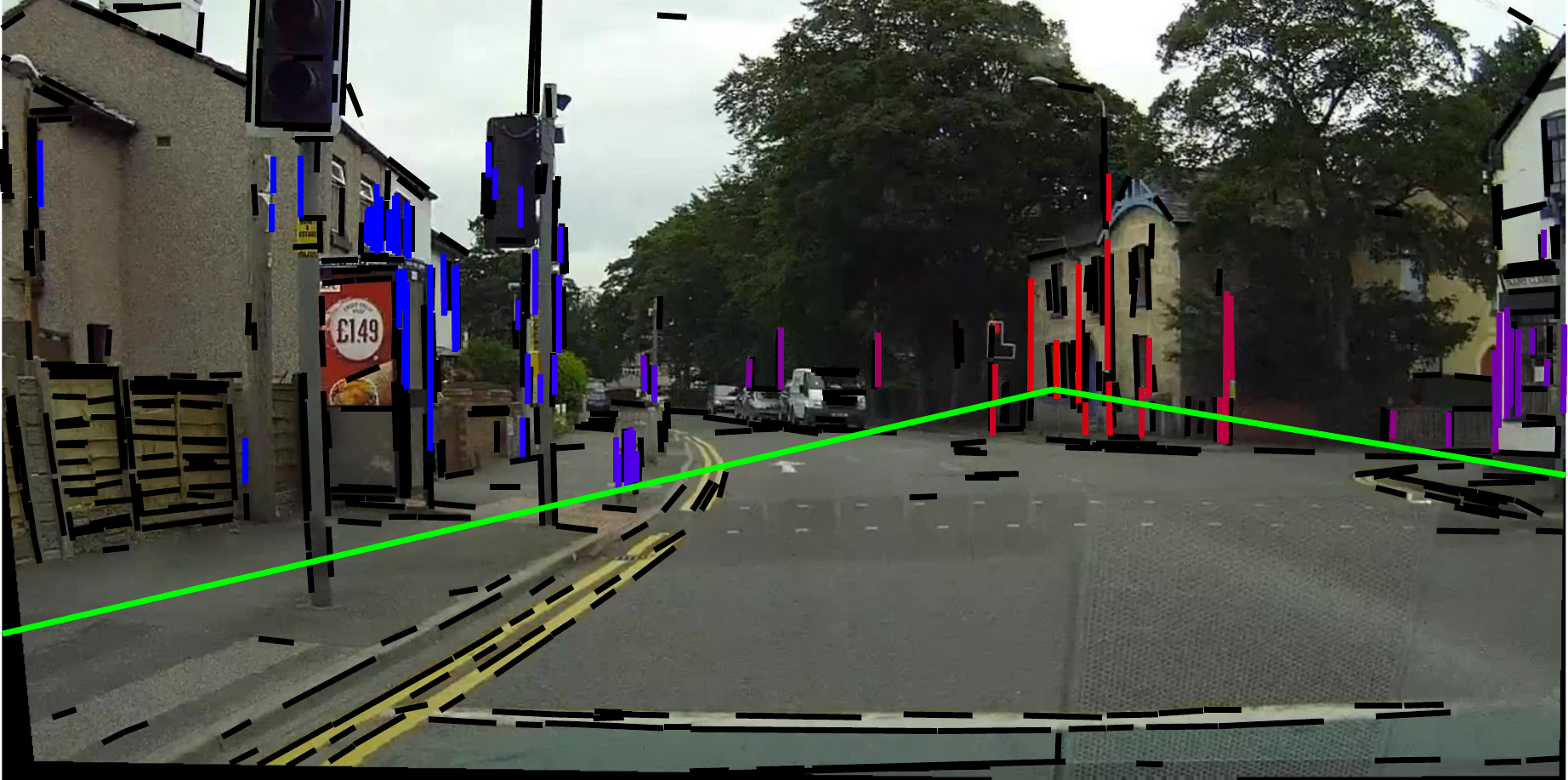} \\ 
\footnotesize (c) A real rolling shutter image & \footnotesize (d) Motion compensated image
\end{tabular} 
\end{center} 
\caption{(a) A synthetic image is generated by simulating the Ackermann motion and (c) a real rolling shutter image captured by GS900 2.7" CMOS car camera. (b) and (d) are motion compensated image and the estimated scene depth by proposed method. Red line segments are at a high depth compared to the blue ones. Black line segments are the  detected outliers, and green lines are the estimated road boundaries.} 
\label{fig:synthetic_tocken}
\end{figure}

In this work, we develop a minimal solver for RS compensation from a \emph{single} distorted image. The monocular RS compensation has been addressed in recent methods~\cite{Su_2015_CVPR,rengarajan2016bows,purkait2017rolling}, however, none of these methods incorporate the translational part of the ego-motion. Moreover, these methods are not fast enough to operate in real time. In contrast, we propose a fast minimal solver for rolling shutter motion  compensation. The proposed algorithm is tailored for the \emph{Ackermann motion}, under the known vertical direction. \emph{This is a common scenario for automotive vehicles moving on a horizontal plane}. In conjunction with {\tt RANSAC},  the proposed solver is executed in real time. In Figure~\ref{fig:synthetic_tocken}, we display our result on a synthetic distorted image\footnote{Image source: Google street view, Cambridge \url{https://www.google.com/streetview/}} and on a real rolling shutter image. The result indicates that the proposed method estimates the rolling shutter camera motion quite accurately along with the image depth and the road boundaries. 

\subsection{Related Work}
\noindent The existing RS compensation methods can be categorized into multi-frame based and single-frame based methods. 
External hardware devices have been utilized~\cite{hee2014gyro,jia2012probabilistic,patron2015spline} to acquire camera motion directly in videos. However, these methods deal with the rotational motion only. The geometry of the scene has also been exploited~\cite{albl2015r6p,Dai_2016_CVPR,forssen2010rectifying,grundmann2012calibration,saurer2013rolling,lovegrove2013spline,forster2017svo} using multiple RS images. These methods detect a number of interest points and track over the frames. The tracked points are then exploited to estimate the camera motion. The camera poses for the other rows are then interpolated in order to compensate the RS motion. Nonetheless, none of these methods can directly be applied to {Single-frame} RS compensation. 

Su~\emph{et al.}~\cite{Su_2015_CVPR} utilize motion blur to extract information about the camera motion from a single RS frame. Rengarajan \emph{et al.}~\cite{rengarajan2016bows} utilize the straightness property of the detected arc segments, the length constancy and the angle constancy to estimate the motion. However,~\cite{rengarajan2016bows,Su_2015_CVPR} do not incorporate the translation motion which cannot be disregarded in the Ackermann motion. Moreover, these methods require a nonlinear optimization which has a high runtime complexity. 

In this work, we estimate the Ackermann motion under the known vertical direction. The proposed method also estimates an approximate scene depth while estimating the motion parameters. Moreover, our method is free from the aforementioned drawbacks. 
The contributions are summarized as follows: 
\begin{itemize}[leftmargin=1em,itemsep=1pt,parsep=1pt] 
\item A minimal solver is developed by utilizing vertical line segments to compensate the rolling shutter camera motion. The proposed algorithm is tailored for the Ackermann motion principle which is the   common scenario for automotive vehicles.  
\item Extensive experiments on simulated data show the effectiveness of the proposed approach---computationally and qualitatively. 
\end{itemize}

\section{Rolling Shutter Camera}
\label{RScamera}
Global shutter and rolling shutter cameras differ in how the image sensors are exposed while capturing the photographs. In the former case, the light strikes at all the rows of the image sensor simultaneously for a constant duration of time. In the latter case, each row is exposed for a regular interval of time, while the camera undergoes a (small) motion. 


Let $\mathbf{P} \in \mathbb{R}^3$ be a 3D point in space w.r.t. the GS camera coordinates. Then the position of the point $\mathbf{P}$ w.r.t. RS camera coordinates [Fig. \ref{fig:ackerman_motion11}(a)] is 
\begin{equation}
\mathbf{P}_{\scA} = {R^t\scA}(\mathbf{P} - {\bT}\scA^t)
\label{eq:3dpoints}
\end{equation}
which can be written in the normalized pixel coordinates as follows \cite{hartley2003multiple, rengarajan2016bows}
\begin{equation}
s^{rs}\scA K\inv{\bp}^{rs} = {R^t\scA}(sK\inv{\bp} - {\bT}\scA^t) , 
\label{eq:original}
\end{equation}
where ${\bp} = [p_1,\;p_2,\;1]^\intercal$ and ${\bp^{rs}} = [p_1^{rs},\;p_2^{rs},\;1]^\intercal$ are the homogeneous pixel coordinates of the pixel $ (p_1,\;p_2)$ of the GS and RS cameras respectively.  $s$ and $s^{rs}\scA$ are  corresponding scene depths. $K$ is the intrinsic camera matrix. $R^t\scA$ and ${\bT}\scA^t$ are the rotation and translation of the vehicle at time $t = \tau p_2^{rs}$ where $\tau$ is the time delay between two successive rows. 
$\cA$ is the Ackermann motion parameters. For readability, in the rest of the paper, we consider ${\bp}$ and ${\bp}^{rs}$ are on the image plane, i.e., pre-multiplied by $K\inv$. Thus \eqref{eq:original} becomes 
\begin{equation}
s{\bp} = s^{rs}\scA{R^t\scA}^\intercal{\bp}^{rs} + {\bT}\scA^t. 
\label{eq:RScamera}
\end{equation}
where {${(R^t\scA)}^\intercal$ is the  transpose of ${R^t\scA}.$} Note that the scene depth $s^{rs}\scA$ varies with the pixels and also with the camera motion. 

\section{Modeling Ackermann motion}
\begin{figure*}
\begin{center}
\begin{tabular}{c@{\hspace{0.2em}}c}
\includegraphics[scale=0.84]{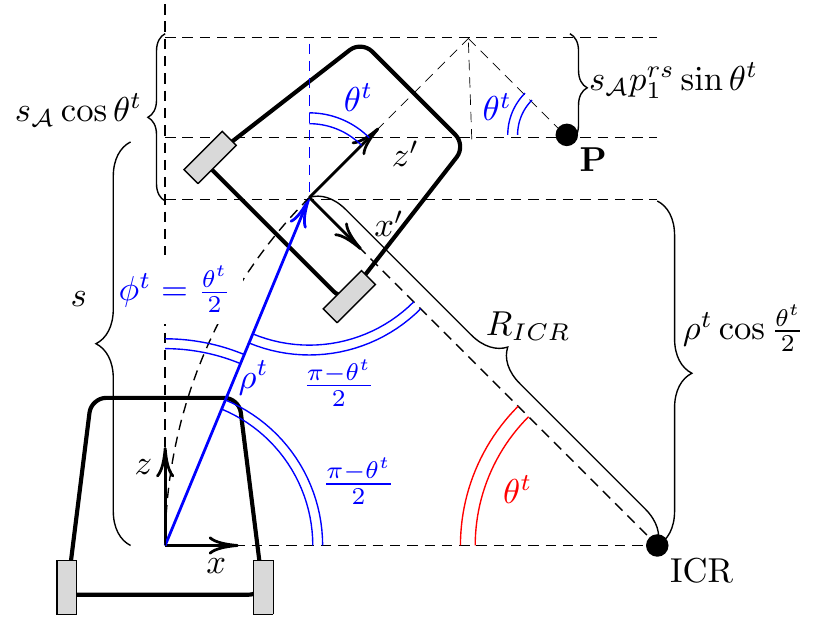} &
\includegraphics[scale=0.92]{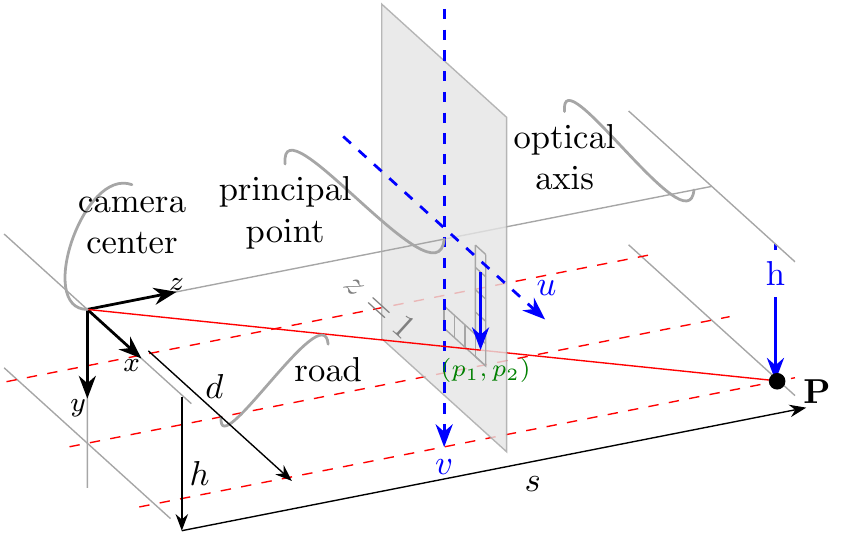}  \\
(a) pure circular motion & (b) pure translational motion 
\end{tabular}
\end{center}
\caption{Ackermann motion: (a) pure circular motion and (b) pure translational motion---it also shows the relation between the scene depth and the pixel coordinates.}
\label{fig:ackerman_motion11}
\end{figure*}

The conventional Ackermann steering principle \cite{scaramuzza2009absolute,scaramuzza2009real,siegwart2011introduction} is to have all the four wheels of a vehicle rolling around a common point during a turn. This principle holds for any automotive vehicle which ensures all the wheels exhibit a rolling motion. This common point is widely known as the Instantaneous Center of rotation (ICR) and is computed by intersecting all the roll axis of the wheels. In Figure~\ref{fig:ackerman_motion11}, we demonstrate such motion.  The radius of the circular motion goes to infinity under a pure forward motion. 

\label{sec:gauge_freedom}

\subsection{Motion model}
\noindent The vehicle is assumed to undertake a fixed translational and angular velocity while capturing the image. If it takes a pure circular motion with a rotation angle $\theta^t$, then, the vehicle must satisfy the circular motion constraints $\phi^t = \frac{\theta^t}{2}$ [Fig. \ref{fig:ackerman_motion11}]. Note that here the $x$-axis is taken along the radius towards the ICR. The relative pose \cite{hee2013motion,hee2014relative} between the vehicles (previous and current  position) can be written as  
\begin{equation}
\begin{array}{l}
{R}^t\scA = 
\arraycolsep=5pt\arraycolsep=5pt\def\arraystretch{1}
\left[ \begin{array}{ccc}
\cos \theta^t & 0 & -\sin \theta^t \\
0 & 1 & 0 \\
\sin \theta^t & 0 & \cos \theta^t
\end{array}
\right], ~~
{\bT}\scA^t = 
\arraycolsep=10pt\def\arraystretch{1} 
\left[ \begin{array}{ccc}
\rho^t\sin \frac{\theta^t}{2}  \\
0 \\
\rho^t\cos \frac{\theta^t}{2} 
\end{array}
\right]
\end{array}
\label{eq:cayleyRot2}
\end{equation} 
where $\theta^t$ is the relative yaw angle and $\rho^t$ is the displacement of the vehicle at time $t$. 
Substituting in \eqref{eq:RScamera}, we obtain the scene depth 
\begin{equation}
s = s^{rs}\scA ( \cos\theta^t - p_1^{rs}\sin\theta^t ) + \rho^t \cos \frac{\theta^t}{2}
\end{equation} 
which can also be verified by Figure~\ref{fig:ackerman_motion11}(a). Let the vehicle undergo a circular motion with an angular velocity $\alpha$ and a translational velocity $\beta$ on the horizontal plane. Under the assumption,  
\begin{equation}
\begin{array}{l}
{R}^t\scA = 
\arraycolsep=3pt\def\arraystretch{1}
\left[ \begin{array}{ccc}
1 - 2{\alpha}^2t^2 & 0 & - 2\alpha t\gamma_t \\
0 & 1 & 0 \\
2\alpha t\gamma_t & 0 & 1 - 2\alpha^2t^2
\end{array}
\right], ~~
{\bT}\scA^t = 
\arraycolsep=3pt\def\arraystretch{1} \beta t
\left[ \begin{array}{ccc}
 \alpha t \\
0 \\
\gamma_t 
\end{array}
\right],
\end{array}
\label{eq:cayleyRot3}
\end{equation} 
where $~\alpha^2t^2 + \gamma_t^2 = 1$ and $\cA = \{\alpha, \beta\}$ is the set of unknown rolling shutter parameters. $\gamma_t$ was introduced to ensure the unit norm. The scene depth $s^{rs}\scA$ is simplified into $s^{rs}\scA = (s - \beta t)/(1 - 2p_1^{rs}\alpha t) $.  Notice that the terms involving the motion parameters $\cA$ of third and higher orders are ignored. Under small rotations, the rotation matrix can be linearly approximated  \cite{albl2015r6p}. Therefore, further dropping the second order terms from \eqref{eq:cayleyRot3} (eventually, $\gamma_t$ becomes $1$) and substituting in \eqref{eq:RScamera}  
\begin{equation}  
\begin{array}{l}
 s\bp = \frac{s - \beta t}{1 - 2p_1^{rs}\alpha t} (I_3 + 2[\br\scA^t]_\times){\bp}^{rs} + \beta t {\bs}\scA^t
\end{array}
\label{eq:simplifiedRot}
\end{equation} 
where $[\cdot]_\times$ denotes the skew-symmetric cross-product matrix and $I_3$ is the $3\times 3$ identity matrix. $\br^t\scA = [0, \alpha t, 0]^\intercal$ and ${\bs}\scA^t = [\alpha t, 0, 1]^\intercal$ are the angular rotation and the contribution of the circular motion to the translation.

\subsection{Minimal Solver with known vertical direction}
\noindent We assume that the vertical direction is known, which is readily available from an external hardware source, e.g., inertial measurement unit (IMU). The accelerometers and gyroscopes of the IMUs provide a precise measure of  the orientation of the gravitational acceleration (roll and pitch angles) in the camera frame \cite{albl2016rolling}, from which one can easily derive the vertical direction $\be_y$. Without loss of generality, we can assume further that the known vertical direction of the camera is also vertical w.r.t. the scene coordinates, i.e., $\be_y = [0,\; 1,\; 0]$. 

Let $\bu_i = [u_{i1}, u_{i2}, u_{i3}]^\intercal$ and $\bv_i = [v_{i1}, v_{i2}, v_{i3}]^\intercal$ be the motion compensated end points \eqref{eq:simplifiedRot} of a line segment $l_i$. Then, the  normal of the interpretation plane of $l_i$ is $ \bu_i \times \bv_i $. If the $l_i$ is vertical, then the  interpretation plane must pass through the vertical axis, which leads to $(\bu_i \times \bv_i)^\intercal \be_y = 0 $, i.e.,
\begin{equation}
u_{i3}v_{i1} - u_{i1}v_{i3} = 0
\label{eq:rollingshutterminimal}
\end{equation}
Substituting in terms of rolling shutter pixel coordinates,  the above \eqref{eq:rollingshutterminimal} leads to a polynomial equation with unknown motion parameters $\cA$ and unknown scene depth $s$.  
Apparently, the rolling shutter compensation requires an estimation of the scene depth $s$ at every individual pixel. 
In the following section, we employ a simplified parametric representation of the scene depth $s$, which is approximately valid for vehicles driving in urban areas.    

\subsection{Parametrization of the depth} 
\begin{figure*}
\begin{center}   
\begin{tabular}{c@{\hspace{0.2em}}c}
\includegraphics[width=0.42\textwidth]{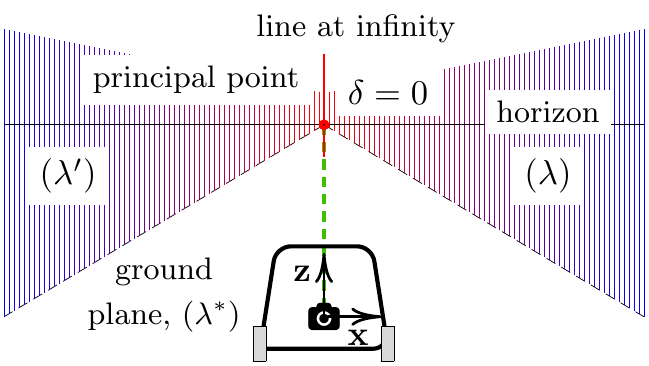} &
\includegraphics[width=0.42\textwidth]{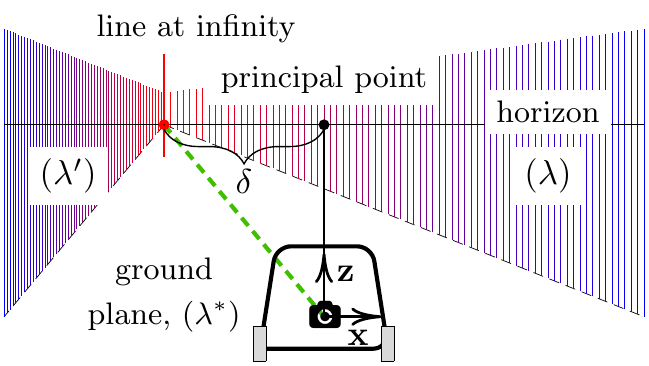} \\ 
(a) Pure translation, facing front & (b) Ackermann motion, facing sideways  
\end{tabular}
\end{center}
\caption{The inverse of the scene depth is assumed to be zero for the column corresponding to the line at infinity and then increases linearly along the sides of the road. The vertical lines corresponding to red lines are at a large depth compared to the blue ones. }
\label{fig:ackerman_motion}
\end{figure*}

\noindent The scene is assumed to be composed of two vertical planes (buildings, etc.) and one horizontal plane (road). In Figure \ref{fig:ackerman_motion}(a)-(b), we illustrate our scene depth assumptions. The two sides of the roads are approximated by two vertical planes which intersect at the line at infinity. The road is considered as the horizontal ground plane. Then the horizon must pass through the principal point of the camera as the camera is assumed to be vertical. The scene depth at any pixel is considered as the minimum depth of the scene surrounded by the planes. 

Let $(p_1,\; p_2)$ be the normalized pixel coordinates of a 3D point $\mathbf{P} \in \mathbb{R}^3$ on the ground plane at a depth $s$. By the pinhole camera model \cite{hartley2003multiple}, $\frac{p_2}{1} = \frac{h}{s}$, i.e., $s\inv = h\inv p_2$ [Fig. \ref{fig:ackerman_motion11}(b)], where $h$ is the height of the camera from the ground plane. Similarly, for the points on any of the vertical planes, $s\inv = d\inv p_1$,  where $d$ is the distance of the vehicle from the vertical plane. In general, for the vehicles facing the vertical planes sideways,  $s\inv = d\inv (p_1 - \delta)$ where $\delta$ is the column corresponds to line at infinity [Fig. \ref{fig:ackerman_motion}(b)]. Thus, the inverse of the scene depth $s\inv$ is linear in column number of the pixel in normalized image coordinates. 

By our assumptions, $s\inv = \max\{\lambda^\prime (p_1 - \delta),\; \lambda (p_1 - \delta),\;  \lambda^\ast p_2\}$ is the inverse depth at the pixel $(p_1,\; p_2)$. The column $\delta$ corresponds to the line at infinity. $\lambda^\prime, \lambda$ are the inverse of the distances of the camera from the vertical planes. $\lambda^\ast$ is the inverse of the height of the camera which is assumed to be known. By the construction, the vertical line segments must lie on any of the vertical planes in 3D represented by $\lambda^\prime, \lambda$. Thus, \eqref{eq:simplifiedRot} becomes 
\begin{equation}  
\begin{array}{l}
 \bp~~ = \frac{(1 - \beta t s\inv)}{(1 - 2p_1^{rs}\alpha t)}(I_3 + 2[\br\scA^t]_\times){\bp}^{rs} + \beta t s\inv {\bs}\scA^t \\
 s\inv = \lambda^\prime[p_1 - \delta]_- + \lambda [p_1 - \delta]_+
\end{array}
\label{eq:simplifiedRot8}
\end{equation} 
where $[x]_- = -\min\{0,\; x\}$  and $[x]_+ = \max\{0,\; x\}$ are non-zeros only on the respective vertical planes. Clearly, the scale of the depth parameter $\lambda^\prime,\lambda$ and the scale of the translational velocity $\beta$ introduce a gauge freedom. Thus, knowing one parameter, other two can be estimated explicitly. In this work, we fix the gauge by choosing $\lambda^\prime = 1$, i.e., assume that the  position of the vehicle w.r.t. one of the vertical plane is known, and the translational velocity $\beta$ of the vehicle and the parameter $\lambda$ of the other side of the vertical plane are estimated. On the other hand, if the velocity of the vehicle $\beta$ is known, estimation of the parameters of the vertical planes $\lambda^\prime$ and $\lambda$, leads to the same solver. Further, if the detected vertical line segments touch the ground plane whose pixel coordinates are known, then the velocity of the vehicle as well as the location can be computed from $\lambda^\ast$. Nevertheless, the distorted image can always be compensated without knowing the scale (velocity, depth etc.).  

To solve two motion parameters $\{\alpha,\beta\}$ and two depth parameter $\{\lambda, \delta \}$, we require four line segments to estimate the model as well as the depth of the scene. Notice that in above, 
$s\inv = [p_1 - \delta]_- + \lambda [p_1 - \delta]_+$ is computed w.r.t. GS camera coordinates. Substituting \eqref{eq:simplifiedRot8} into \eqref{eq:rollingshutterminimal} leads to a system of  eight polynomial equations of degree five with $29$ monomials.  The Gr\"obner basis solution \cite{bujnak2008general,kukelova2008automatic} for this system involves eliminating a $467 \times 478$ matrix, which takes $\approx 0.1$ second for each problem instance. Note that in conjunction with {\tt RANSAC}, the whole procedure for robustly estimating the model parameters would take $0.1N$ where $N$ is the number of {\tt RANSAC} iterations. 

The efficiency is improved with the following approximations. The inverse of depth $s\inv$ is assumed to be linear w.r.t. RS pixel coordinates, i.e., $ s\inv = [p_1^{rs} - \delta]_- + \lambda [p_1^{rs} - \delta]_+$. The experiments  demonstrate that even with such approximation, we can still estimate the parameters very accurately. The problem is further simplified by considering three line segments in one of the vertical planes which are solved by Gr\"obner basis method. It computes $\alpha, \beta$ and $\delta$.  These are then substituted in the constraint  \eqref{eq:rollingshutterminimal} of the other line segment---forms a quadratic equation in $\lambda$ of which least absolute solution is chosen. 
The Gr\"obner basis method generates an elimination template matrix of size $12\times 15$ on which a G-J elimination method is performed. It produces $3$ solutions and takes $\approx 0.001s$ for each problem instance. Among the solutions, only the potential ones are considered by discarding unrealistic solutions (e.g., solutions corresponding to the absolute translation velocity above $200$km/h and absolute angular velocity over $90$deg/s). 

In the following, we furnish the pseudo-code of the proposed algorithm (Algo.~\ref{algo:RS}). Note that proposed algorithm is a minimal solver that can estimate the rolling shutter motion robustly in conjunction with {\tt RANSAC} (described in Algo.~\ref{algo:RANSAC}). An interested reader can continue to the following sections else  can move safely to the experiment section (sec. \ref{sec:exp}). 

\begin{algorithm}
    \caption{Rolling Shutter Motion Compensation}
    \begin{algorithmic}[1]
        \Procedure{CompensateMotion}{~Input image~}
        \State {\tt detect line segments} $l_i:=({\bu}^{rs}, {\bv}^{rs})$
        \State {\tt prune line segments} ~~~~~~~\emph{/*  see sec.} \ref{sec:paraset} \emph{*/} 
        \State {\tt estimate}$(\alpha, \beta)$ {\tt and depth}$(\delta, \lambda)$ {\tt algo.\ref{algo:RANSAC}} 
        \State {\tt warp compensated image} ~\emph{/* see sec.} \ref{sec:imgrect} \emph{*/} 
        \State \textbf{return} {~\tt motion compensated image}  
        \EndProcedure
    \end{algorithmic}
\label{algo:RS}
\end{algorithm}

\begin{algorithm}
    \caption{{\tt RANSAC} robust estimation of $(\alpha, \beta, \delta, \lambda)$}
    \begin{algorithmic}[1]
        \Procedure{RansacAckermann}{$l_i:=({\bu}^{rs}, {\bv}^{rs})$}
        \While{{\tt count} $\not=$ {\tt maximum iterations}}
            \State $({\bu}^{rs}, {\bv}^{rs})\gets$ {\tt get 4 random} $({\bu}^{rs}, {\bv}^{rs})$ \label{4la1}
            \State $({\bu}^{rs}, {\bv}^{rs})\gets$ {\tt pick 3 leftmost} $({\bu}^{rs}, {\bv}^{rs})$
            \State $(\alpha, \beta, \delta) \gets$ {\tt Gr\"obner Basis} $({\bu}^{rs}, {\bv}^{rs})$
            \State $\lambda \gets$ {\tt solve quadratic rightmost} 
            \NoNumber ~~~\label{4la2}
             ~~~~~~~~~~~~~~~~~~~~~~~ \emph{/* substituting $(\alpha, \beta, \delta)$ in  \eqref{eq:rollingshutterminimal} */ } 
            \If {{\tt out of range} $(\alpha, \beta, \delta, \lambda)$} 
            \State \textbf{continue}  ~~~~~~~~~~~~~~~~~~~~~~~~~~~~~~\emph{/* see sec.} \ref{sec:syn} \emph{*/ } 
            \EndIf
            \State {\tt count inliers} $(\alpha, \beta, \delta, \lambda)$ 
            \State {\tt update best-found-so-far} $(\alpha, \beta, \delta, \lambda)$
            \State {\tt count} $\gets$ {\tt count} $+1$
        \EndWhile
        \State \textbf{return} {~\tt best-found-so-far} $(\alpha, \beta, \delta, \lambda)$
        \EndProcedure
    \end{algorithmic}
\label{algo:RANSAC}
\end{algorithm}

\subsection{A more general case} 
\noindent Let us consider a more general case where the camera is installed with a known angle with vertical given by an orthogonal matrix $R_\omega \in$ SO(3). 
Then the pixel coordinate of a 3D point $\mathbf{P} \in \mathbb{R}^3$ in \eqref{eq:3dpoints} w.r.t. the GS and RS camera coordinates become $s^\prime {R_\omega} {\bp} = \mathbf{P} $ and $s\scA^\prime {R_\omega} \mathbf{p}^{rs} = \mathbf{P\scA}$ where $s^\prime$ and $s^\prime{\scA}$ are proportional to the scene depth. Thus, \eqref{eq:3dpoints} becomes  
\begin{equation}
s\scA^\prime {R_\omega}{\bp}^{rs}  = {R^t\scA} \big(\mathbf{P} -  {\bT}\scA^t\big). 
\label{eq:RScamera32}
\end{equation}
The above could also be written in terms of GS co-ordinate frame as follows 
\begin{equation}
s^\prime {R_\omega}{\bp} = s^\prime\scA(I_3 + 2[\br\scA^t]_\times){R_\omega}{\bp}^{rs} + {\bs}\scA^t. 
\label{eq:RScamera33}
\end{equation}
where $\br^t\scA = [0, \alpha t, 0]^\intercal$ is the angular rotation and ${\bs}\scA^t = [\alpha t, 0, 1]^\intercal$ is the contribution of the circular motion to the translation. Note that $(I_3 + 2[\br\scA^t]_\times){R_\omega}$ could also be observed as the linearization of the rotation matrix along $R_\omega$ where $s^\prime = s/( {R_\omega^3}^\intercal \bp )$, $s^\prime\scA = s^{rs}\scA/( {R_\omega^3}^\intercal \bp^{rs})$, $s$ is the scene depth corresponding to the pixel $\bp$ and $R_\omega^3$ is the third row of $R_\omega$.  
Substituting, ${\bp^{rs}}^\prime = \frac{1}{ {R_\omega^3}^\intercal \bp} {R_\omega}{\bp}^{rs}$ and $\bp^\prime = \frac{1}{ {R_\omega^3}^\intercal \bp }{R_\omega}{\bp}$ along with $s^{rs}\scA = (s - \beta t)/(1 - 2{p_1^{rs}}^\prime\alpha t) $ in above, however, we reach again to the similar set of equations as follows 
\begin{equation}  
\begin{array}{l}
 \bp^\prime~~ = \frac{(1 - \beta t {s^\prime}\inv)}{(1 - 2{p_1^{rs}}^\prime\alpha t)}(I_3 + 2[\br\scA^t]_\times){\bp^{rs}}^\prime + \beta t {s^\prime}\inv {\bs}\scA^t \\
 {s^\prime}\inv = \lambda^\prime[p_1^\prime - \delta]_- + \lambda [p_1^\prime - \delta]_+
\end{array}
\label{eq:simplifiedRot101}
\end{equation} 
where $t$ could be replaced by $p_2^{rs}$. Note that substituting above in \eqref{eq:rollingshutterminimal}, we arrive at a set of equations of degree five and solve them by Gr\"obner basis method. It generates an elimination template matrix of size $37\times 43$ on which a G-J elimination method is performed. It produces $6$ solutions for each problem instance which were fed into the {\tt RANSAC} iterations.   

\subsection{Specific Motions: } 

\paragraph{Pure translation motion}
 This is a very common scenario for which the vehicle undergoes a pure translation motion, \ie, $\alpha = 0$. Thus, for the left side of the plane,  \eqref{eq:simplifiedRot8} becomes 
  \begin{equation}
   \bp~~ = \frac{\big(1 - \beta p_2^{rs} (p_1 - \delta))}{(1 - 2p_1^{rs}\alpha t\big)}{\bp}^{rs} + \beta p_2^{rs} (p_1 - \delta) {\bs}\scA^t
  \end{equation}
where ${\bs}\scA^t = [0, 0, 1]^\intercal$. In this scenario, a 3-line algorithm will be sufficient to have a finite solution---two of the line segments lie on one plane and the other line segment on the other plane. We can directly compute $p_1 - \delta$ in terms of GS pixel coordinates as follows
  \begin{align}
    p_1 &= p_1^{rs}\big(1 + \beta \delta p_2^{rs}\big)/\big(1 + \beta p_1^{rs} p_2^{rs}\big) \\ 
   \text{i.e.,~} p_1 - \delta &= \big(p_1^{rs} - \delta \big)/\big(1 + \beta p_1^{rs} p_2^{rs}\big)
  \end{align}
Substituting the pixels in terms of rolling shutter pixel coordinates in \eqref{eq:rollingshutterminimal} and then simplifying the equations, we obtain 
\begin{equation}
\big(v_{i1}^{rs}{u_{i1}^{rs}}u_{i2}^{rs} - u_{i1}^{rs}{v_{i1}^{rs}}v_{i2}^{rs}\big)\beta - \big(u_{i1}^{rs}u_{i2}^{rs} - v_{i1}^{rs}v_{i2}^{rs}\big)\beta \delta = u_{i1}^{rs} - v_{i2}^{rs}
\label{eq:simplifiedRot10}
\end{equation}
Note that $u_{i3}^{rs} = 1$ and $v_{i3}^{rs} = 1$ were substituted during the simplification. Two such constraints \eqref{eq:simplifiedRot10} for two line  segments lead to a unique solution. Thus, three line segments are sufficient to have a finite solution for both of the vertical planes---two of the line segments lie on one plane from which we uniquely compute $\beta$ and $\delta$.  The substitution of the estimated parameters to the other constraint, corresponding to the other line segment, leads to a linear constraint in $\lambda$ which again has a unique solution. Since this closed form solution is obtained without an elimination template, it leads to an extremely fast solver. 
\begin{figure*}
\centering \small   
\begin{tabular}{c@{\hspace{-1em}}c@{\hspace{-1em}}c@{\hspace{-1em}}c} 
{\includegraphics[width=0.26\textwidth]{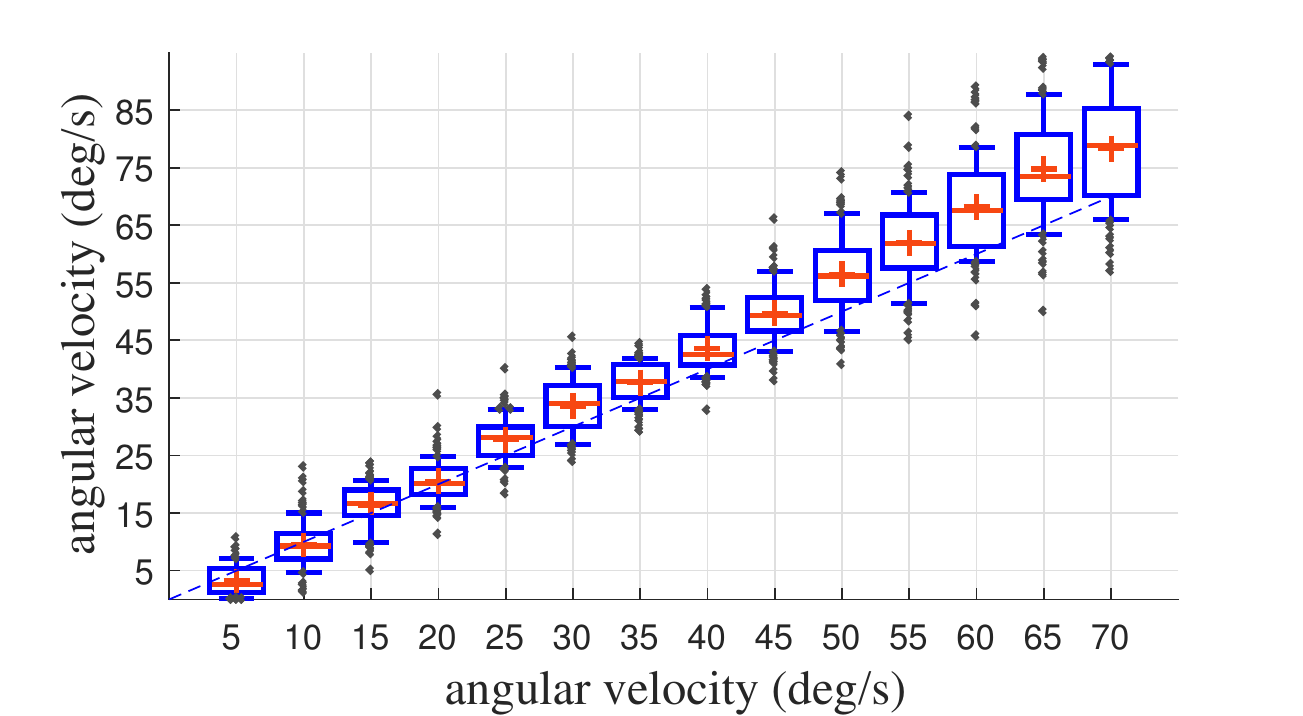}} &
{\includegraphics[width=0.26\textwidth]{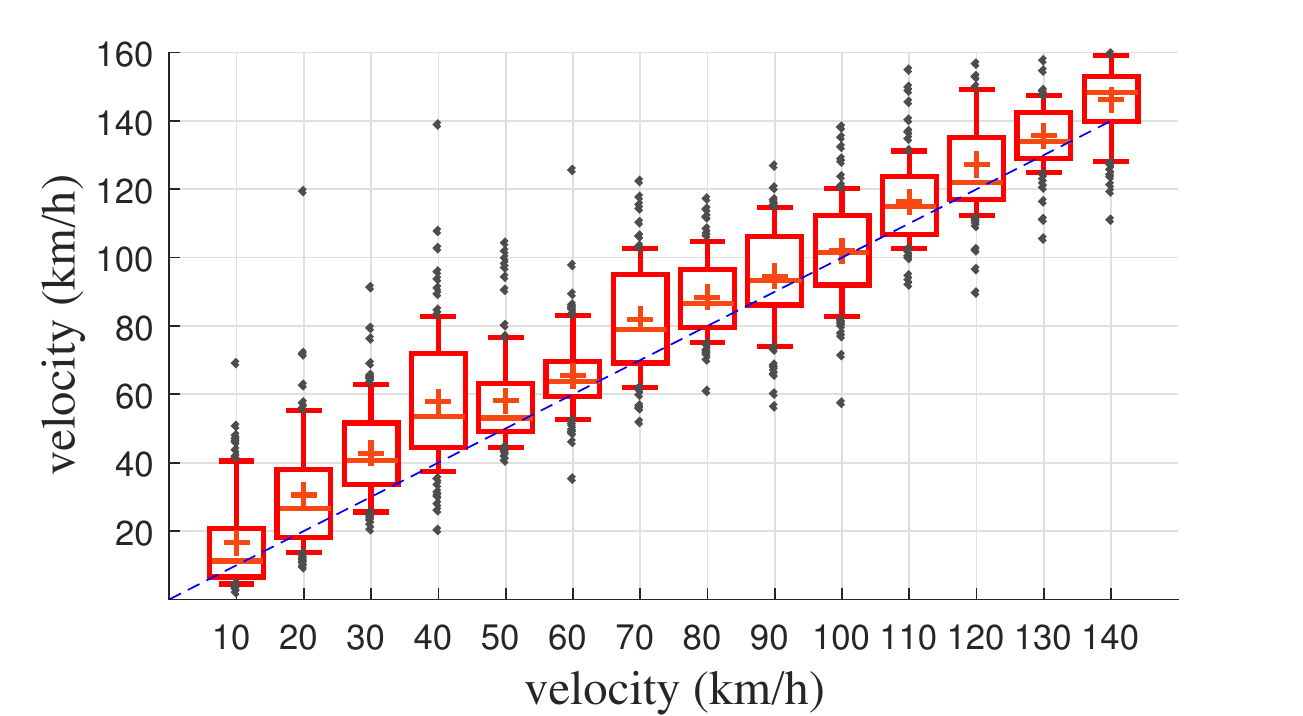}} &
{\includegraphics[width=0.26\textwidth]{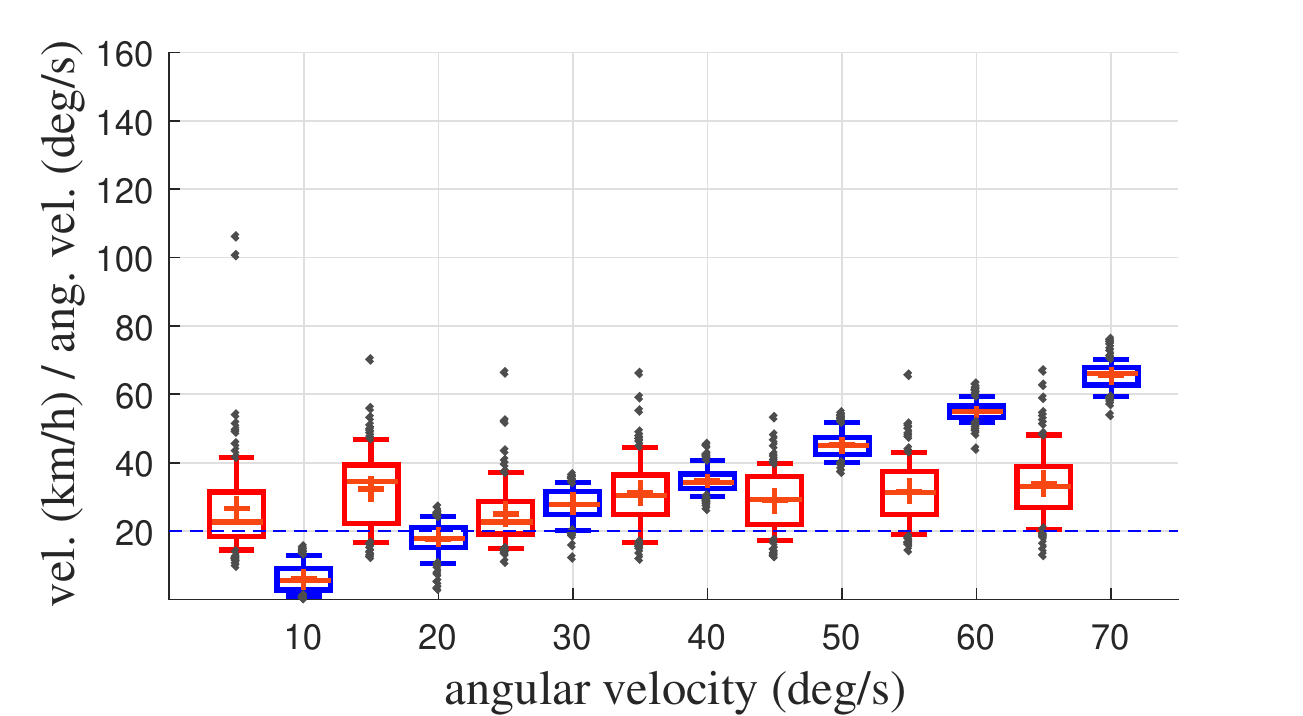}} & 
{\includegraphics[width=0.26\textwidth]{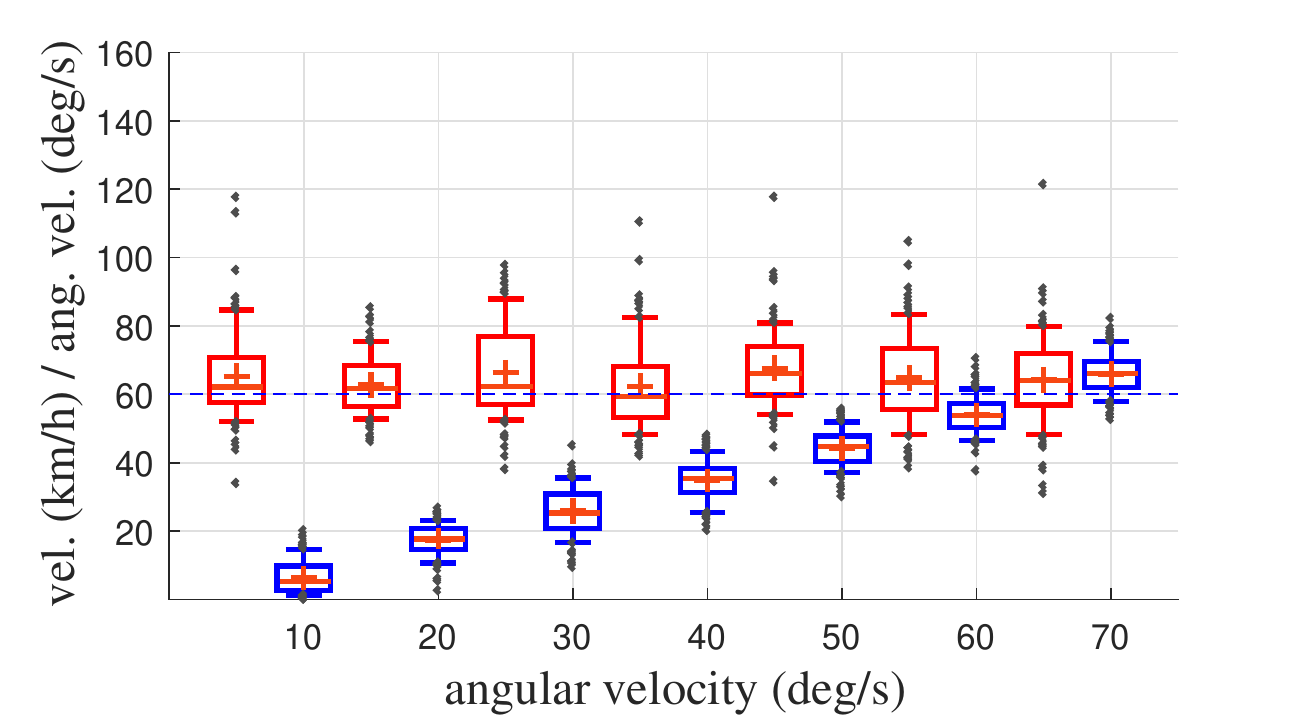}} \\
 (a) {\bf 1-LA}, ang. velo. &  (b) {\bf 3-LA}, trans. velo. &  (c) {\bf 4-LA}, trans. velo. $20$km/h & (d) {\bf 4-LA}, trans. velo. $60$km/h \\ 
{\includegraphics[width=0.26\textwidth]{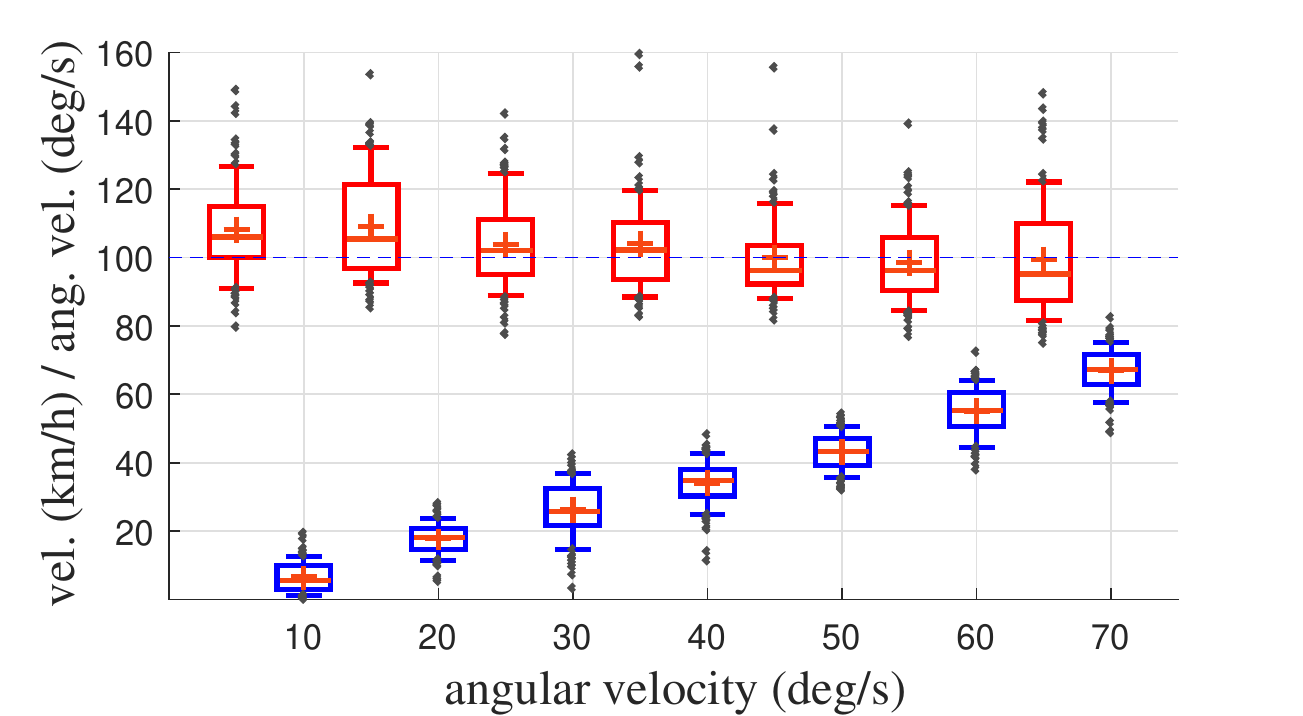}} & 
{\includegraphics[width=0.26\textwidth]{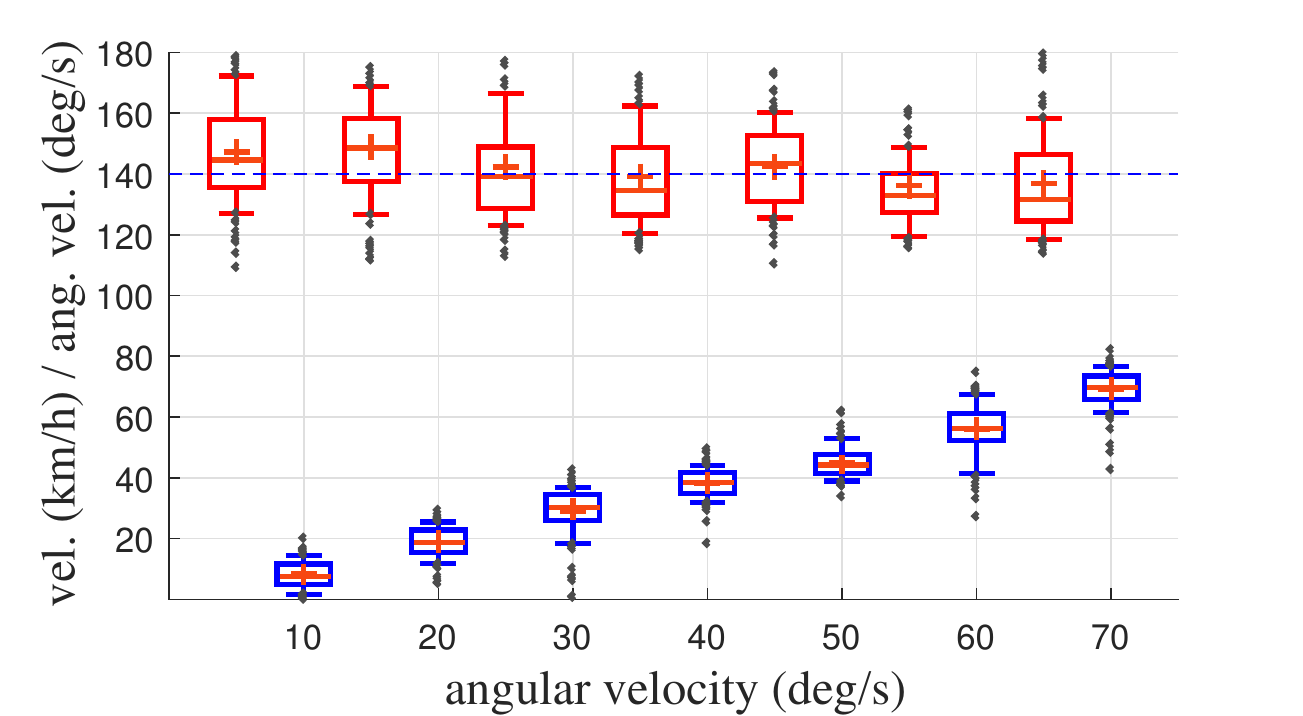}} & 
{\includegraphics[width=0.26\textwidth]{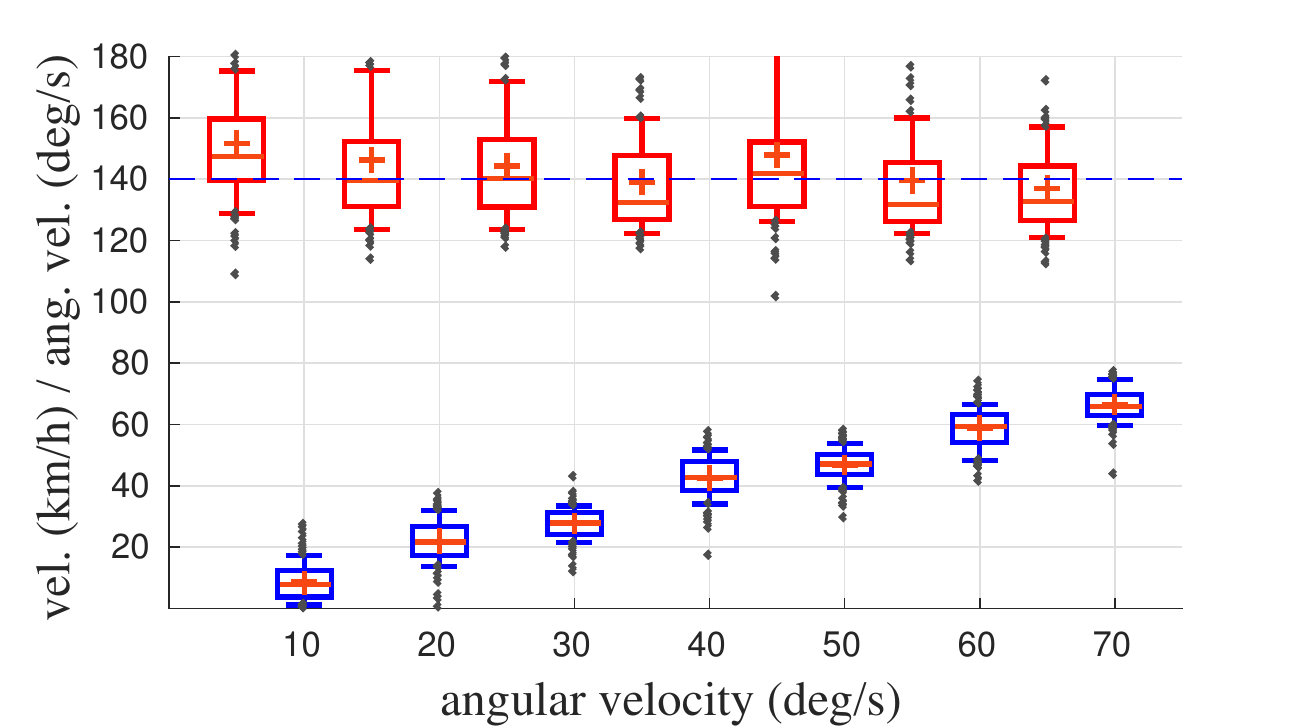}} & 
{\includegraphics[width=0.26\textwidth]{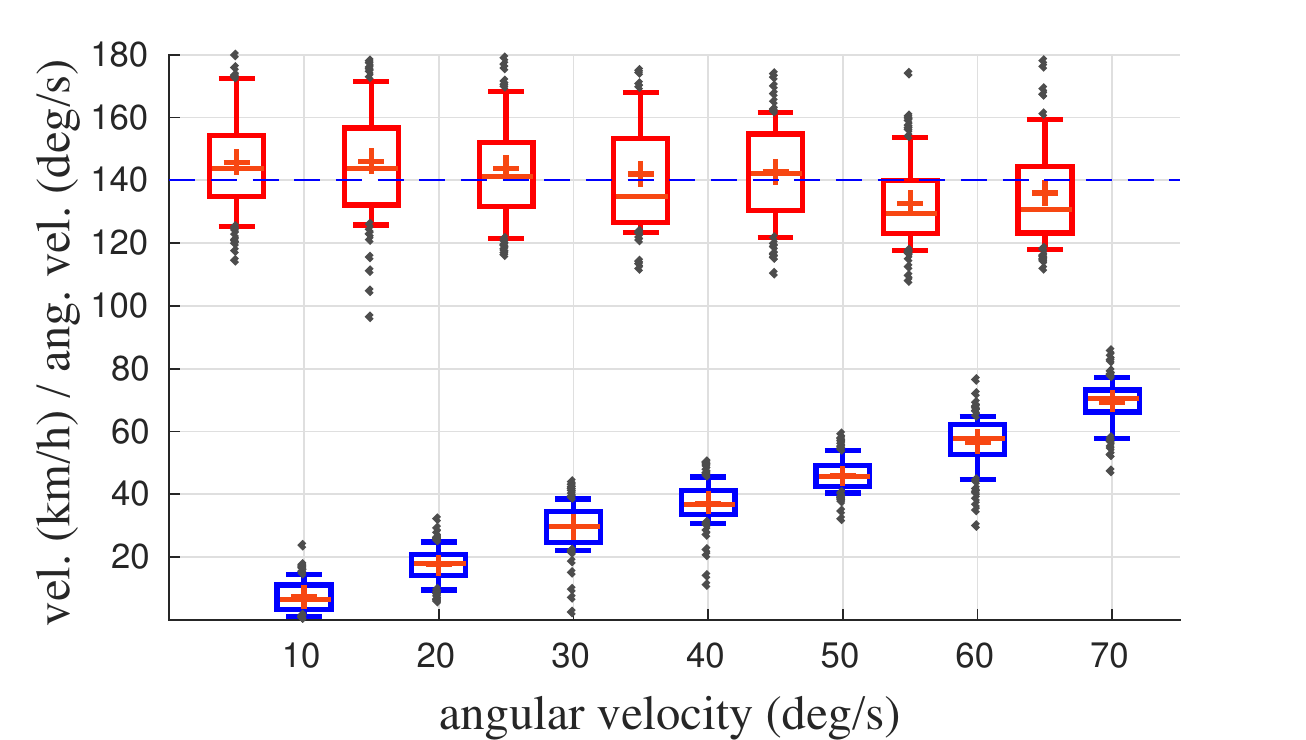}} \\ 
    (e) {\bf 4-LA}, trans. velo. $100$km/h    &  (f) {\bf 4-LA}, trans. velo. $140$km/h &  (g) {\bf 4-LA}, IMU reading error $1\deg$ &  (h) {\bf 4-LA}, depth error $10$cm  
\end{tabular}
 \caption{ (a)-(f) Evaluation of the proposed method {\bf 1-LA}, {\bf 3-LA} and {\bf 4-LA} on a synthetic data. (g) Results under IMU reading error with normal distribution for standard deviation $1$deg~\cite{kukelova2010closed} (vertical assumption) and (h) results under depth error $10$cm (depth assumption). The blue dotted lines represent the simulated velocity. Along $x$-axis and $y$-axis, we display the simulated and the predicted motion. The estimated translational and the angular velocities are represented by red and blue boxes respectively, where the $75\%$ of estimated values lie within the box.  $'+'$ and $'-'$  are the mean and the median of the estimated motions.} 
  \label{fig:SyntheticMotionEstn2}
\end{figure*}

\begin{figure*}[!ht]
\begin{center}
\small 
\begin{tabular}{c@{\hspace{0.05em}}c@{\hspace{0.1em}}c@{\hspace{0.1em}}c}
\includegraphics[width=0.24\textwidth]{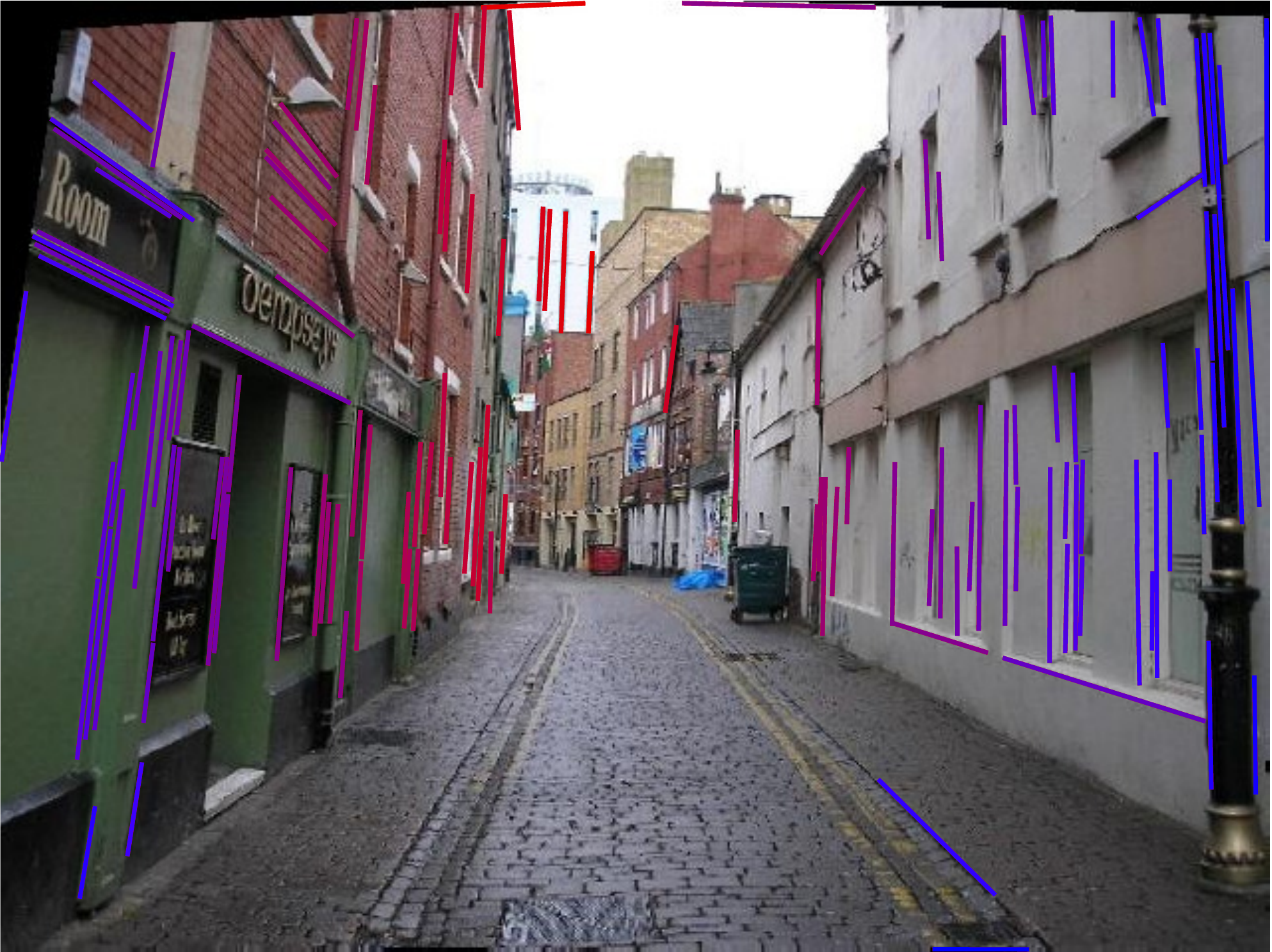} &
\includegraphics[width=0.24\textwidth]{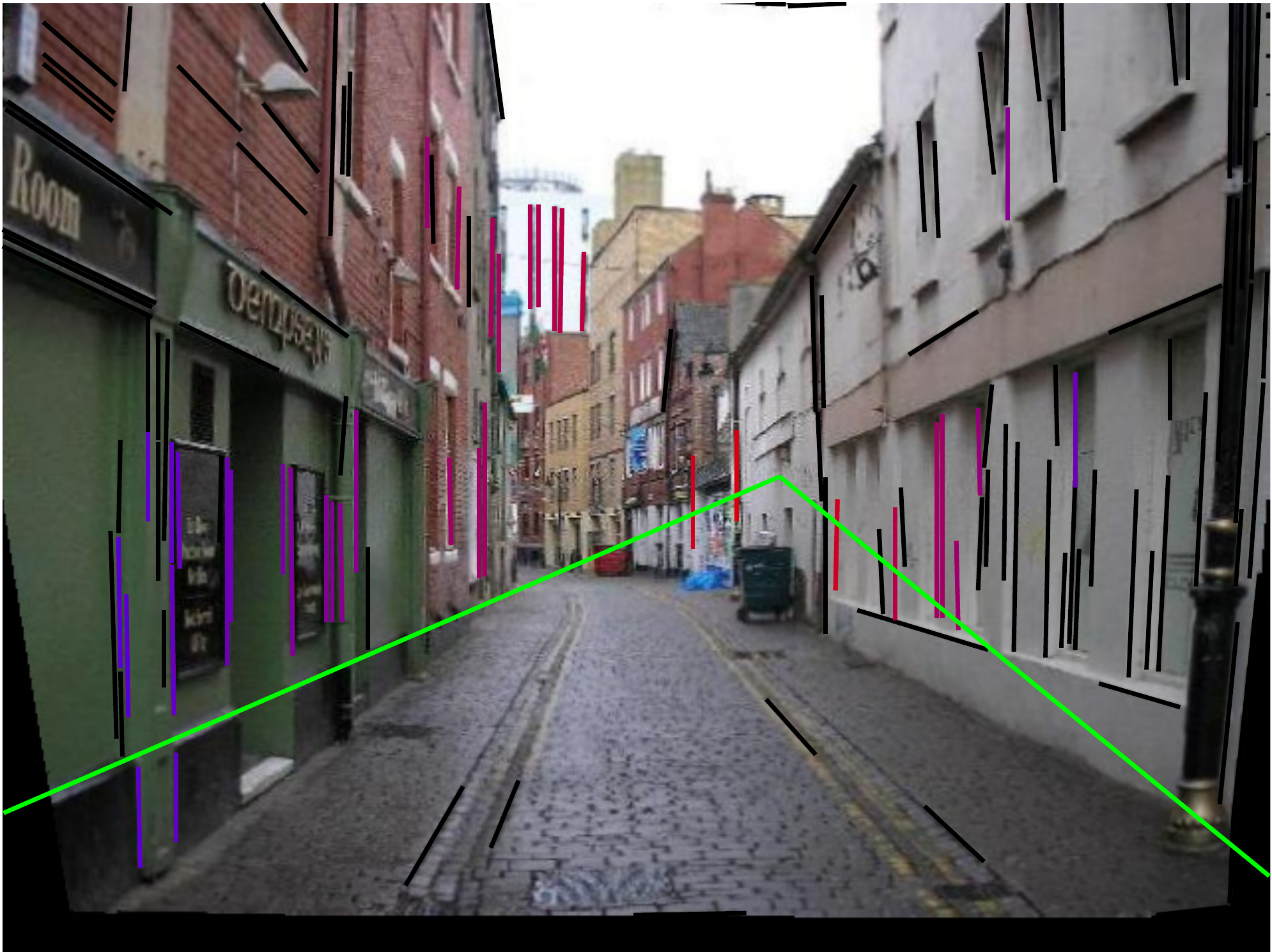} & 
\includegraphics[width=0.24\textwidth]{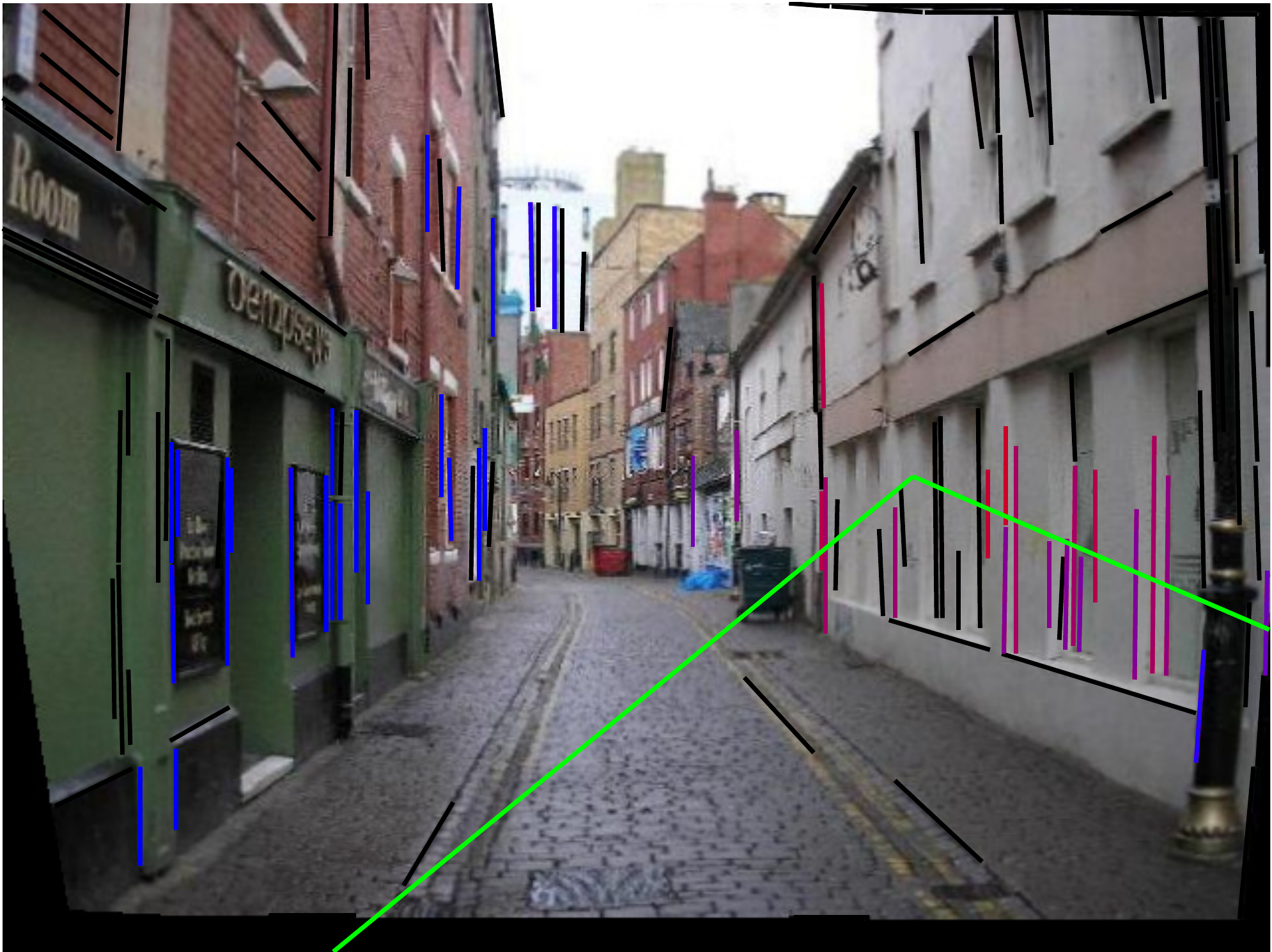} &
\includegraphics[width=0.24\textwidth]{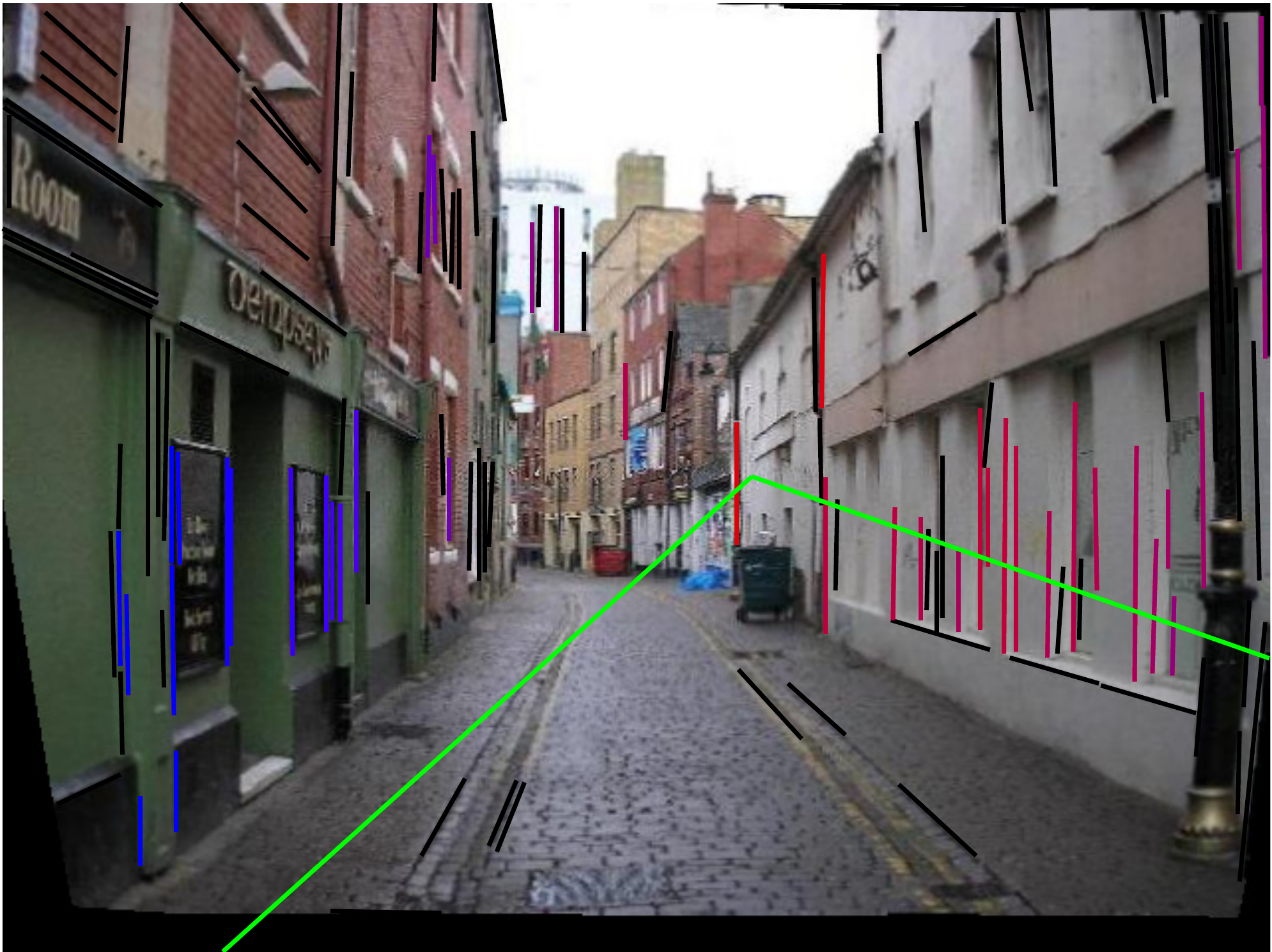} \\ 
 (a) rolling shutter image--- & (b) compensated image--- & (c) compensated image--- & (d) compensated image--- \\ 
 angular velocity $40$deg/s, &  angular velocity $40$deg/s, &  angular velocity $36$deg/s, &  angular velocity $25$deg/s, \\ 
 translational velocity $60$km/h, &  translational velocity $84$km/h, &  translational velocity $37$km/h, &  translational velocity $51$km/h, \\ 
 avg. disp. length $4.72$pixels  &  avg. disp. length $0.88$pixels  &  avg. disp. length $1.24$pixels  &  avg. disp. length $1.53$pixels  
\end{tabular}
\end{center}
\caption{Three different extreme results (executed three times) on one of the instances of the synthetic data of Figure \ref{fig:SyntheticMotionEstn2}. Although, the estimated angular velocities and the translational velocities are not very accurate in these chosen  cases, proposed method still reduces the pixel displacements occurred during rolling shutter motion. The average pixel displacement is computed by robust matching of patches from the original image and the compensated image.  } 
\label{fig:synthetic_data} 
\end{figure*}

\begin{figure}
\begin{center} 
\small 
\begin{tabular}{c@{\hspace{0.05em}}c} 
\includegraphics[width=0.23\textwidth]{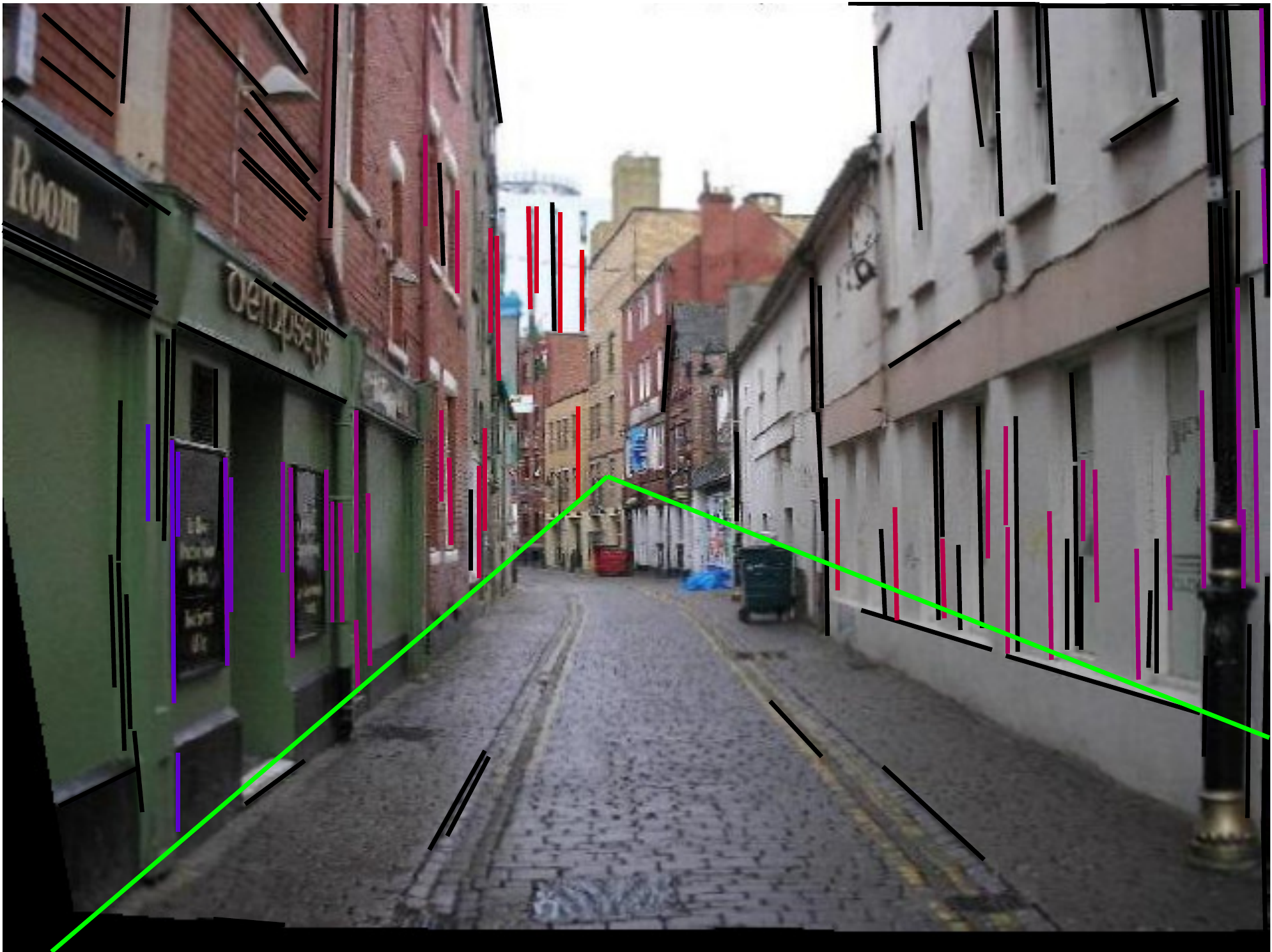} & 
\includegraphics[width=0.23\textwidth]{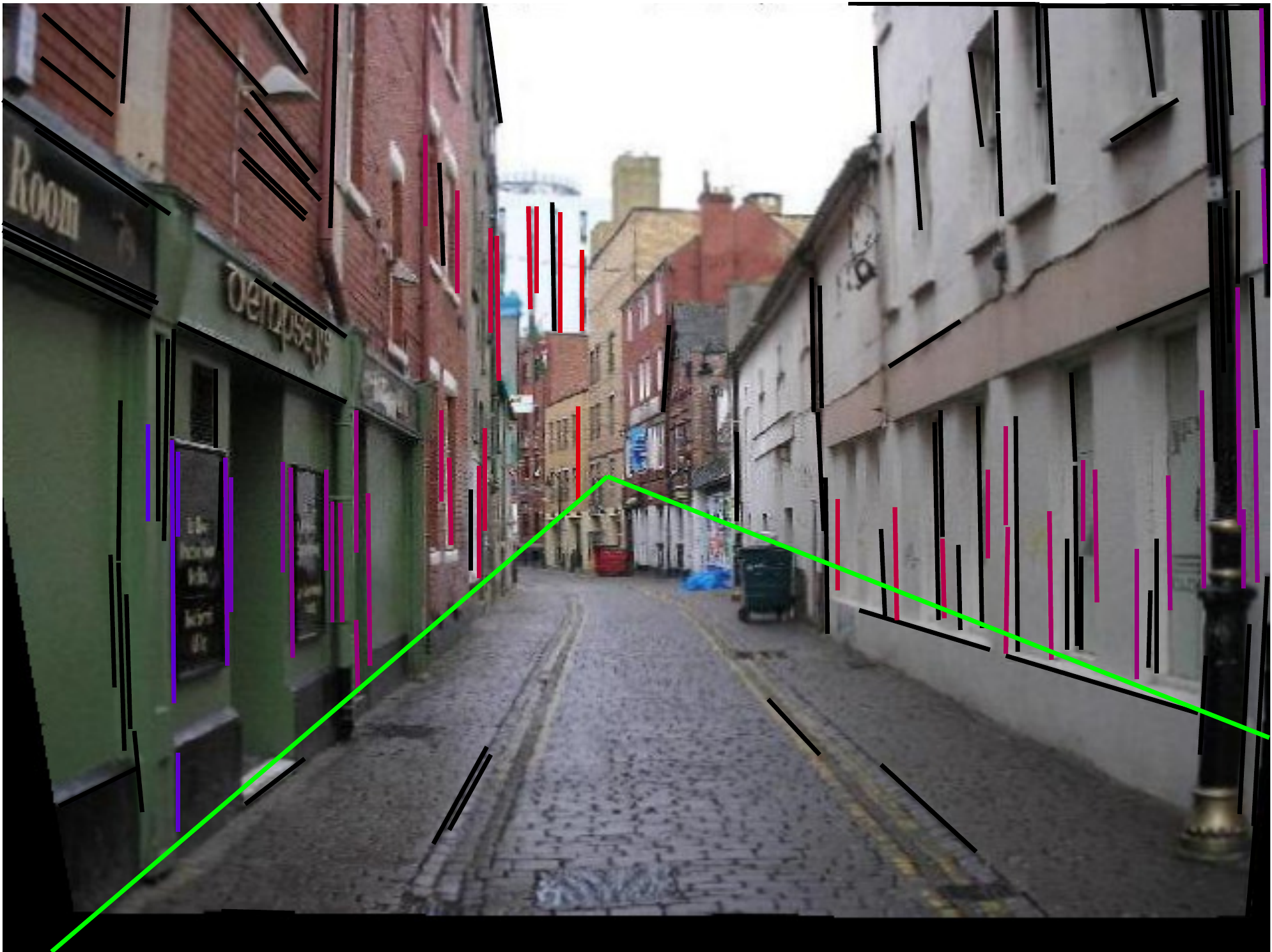} \\ 
(a) compensated image--- & (b) compensated image--- \\ 
 avg. disp. length $0.80$pixels  &  avg. disp. length $1.08$pixels  
\end{tabular}
\end{center}
\caption{ (a) A random instance of Figure \ref{fig:SyntheticMotionEstn2} with estimated translational velocity $59$km/h and the angular velocity $43$deg/s, and (b) under the estimation error of the height of the camera as  $10$cm. We observe very small image quality degradation under the violation of the assumption.} 
\label{fig:synthetic_data22} 
\end{figure} 

\paragraph{Pure rotational motion}
If the vehicle undergoes a pure rotational motion, i.e., $\beta = 0$. The change of depth $s^{rs}\scA $ will not affect the vertical direction \eqref{eq:rollingshutterminimal}. In this scenario, the geometric relation between the GS and RS image frames is become 
\begin{equation}  
\begin{array}{l}
 \bp \propto (I_3 + [\br\scA^t]_\times){\bp}^{rs} 
\end{array}
\label{eq:simplifiedRot3}
\end{equation} 
where  $\br^t\scA = [0, \alpha t, 0]^\intercal$.  After simplification, the  constraint \eqref{eq:rollingshutterminimal} is simplified to   
\begin{equation}  
\begin{array}{l}
\big({u_{i1}^{rs}} u_{i2}^{rs} v_{i2}^{rs} - {v_{i1}^{rs}} v_{i2}^{rs} u_{i2}^{rs}\big)\alpha^2 - \\ \big(u_{i1}^{rs}v_{i1}^{rs}v_{i2}^{rs} - v_{i1}^{rs}u_{i1}^{rs}u_{i2}^{rs}  + u_{i2}^{rs} - v_{i2}^{rs}\big)\alpha + u_{i1}^{rs} - v_{i1}^{rs} = 0 
\end{array}
\label{eq:simplifiedRot5}
\end{equation} 
which is a quadratic in $\alpha$ and only the solution with the least absolute value is considered. Thus, a single line segment is sufficient to estimate $\alpha$, this leads to an efficient and a much faster 1-line algorithm. However, this scenario is not an Ackermann motion (also different from motion demonstrated in Figure \ref{fig:ackerman_motion11}(a)) and the solution is presented here for completeness. Note that the existing accurate nonlinear optimization techniques, e.g., \cite{rengarajan2016bows} can be utilized in this case which is, nevertheless, much slower than the proposed minimal solver.

\begin{figure*}
\centering 
\begin{tabular}{l@{\hspace{1.3em}}c@{\hspace{0.1em}}c@{\hspace{0.1em}}c@{\hspace{0.1em}}c@{\hspace{0.1em}}c} 
\begin{picture}(1,25)
  \put(0,0){\rotatebox{90}{~~~~~\color{blue}{Input}}}
\end{picture} & 
{\includegraphics[width=0.2\textwidth]{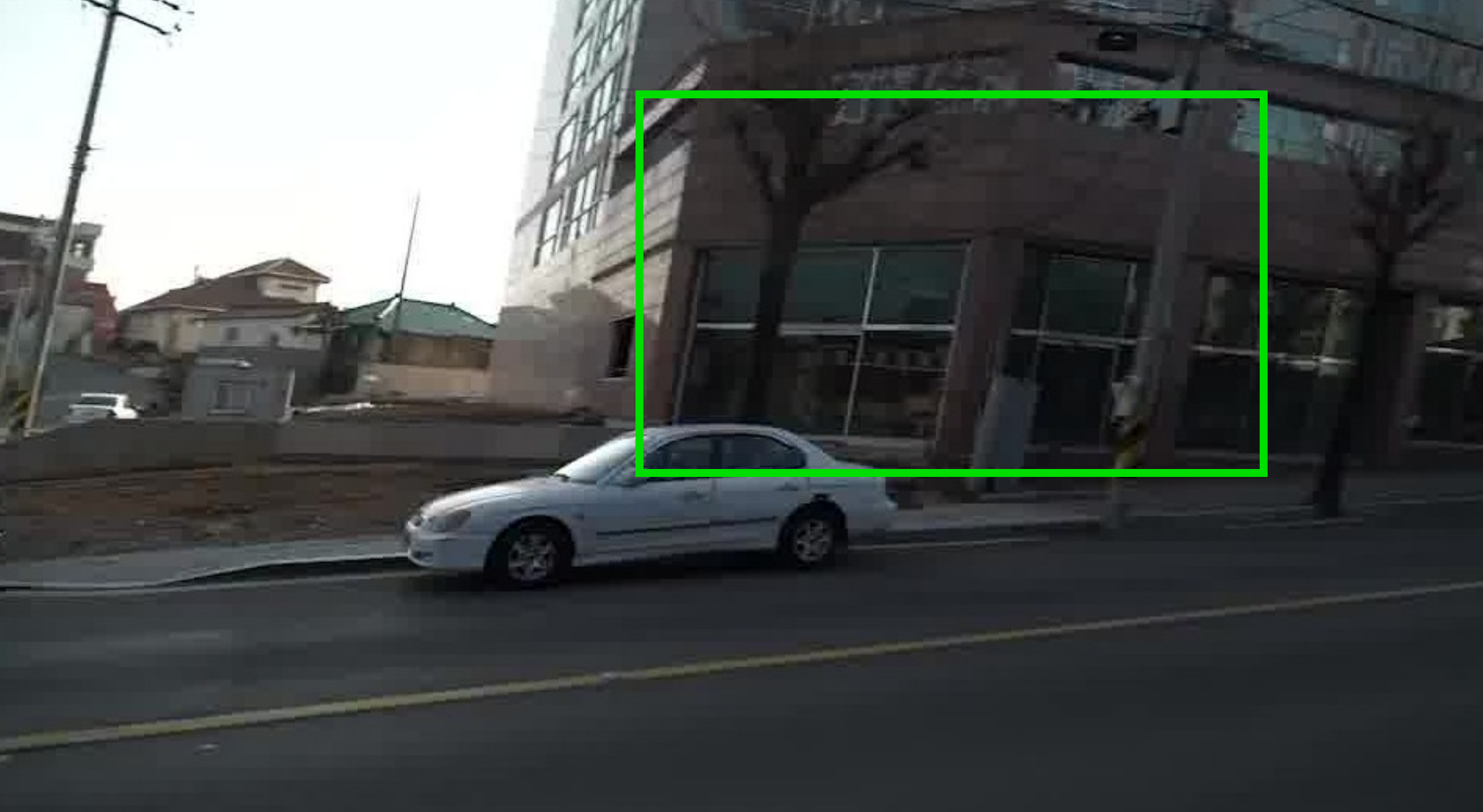}} &
{\includegraphics[width=0.22\textwidth]{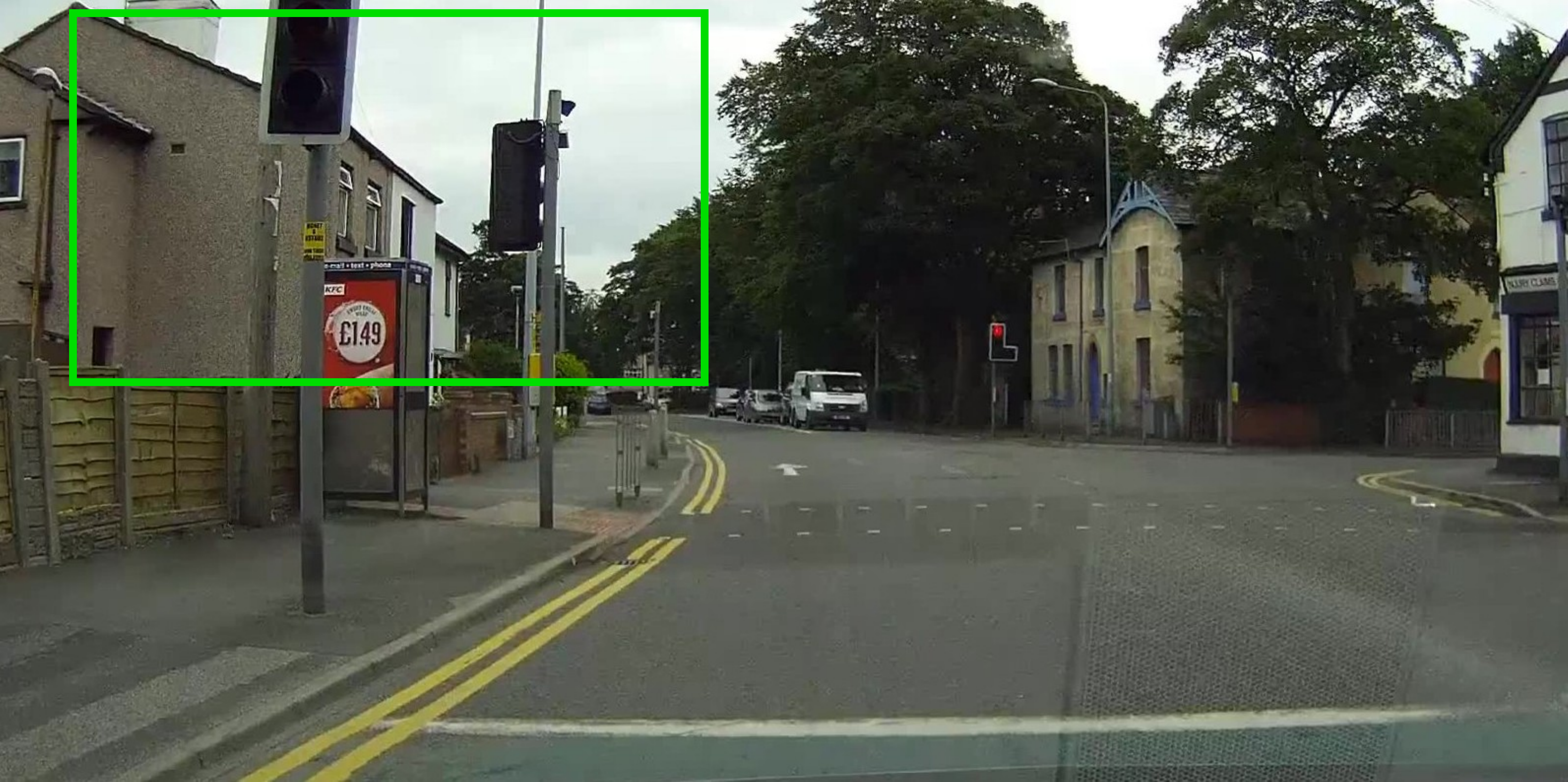}} & {\includegraphics[width=0.2\textwidth]{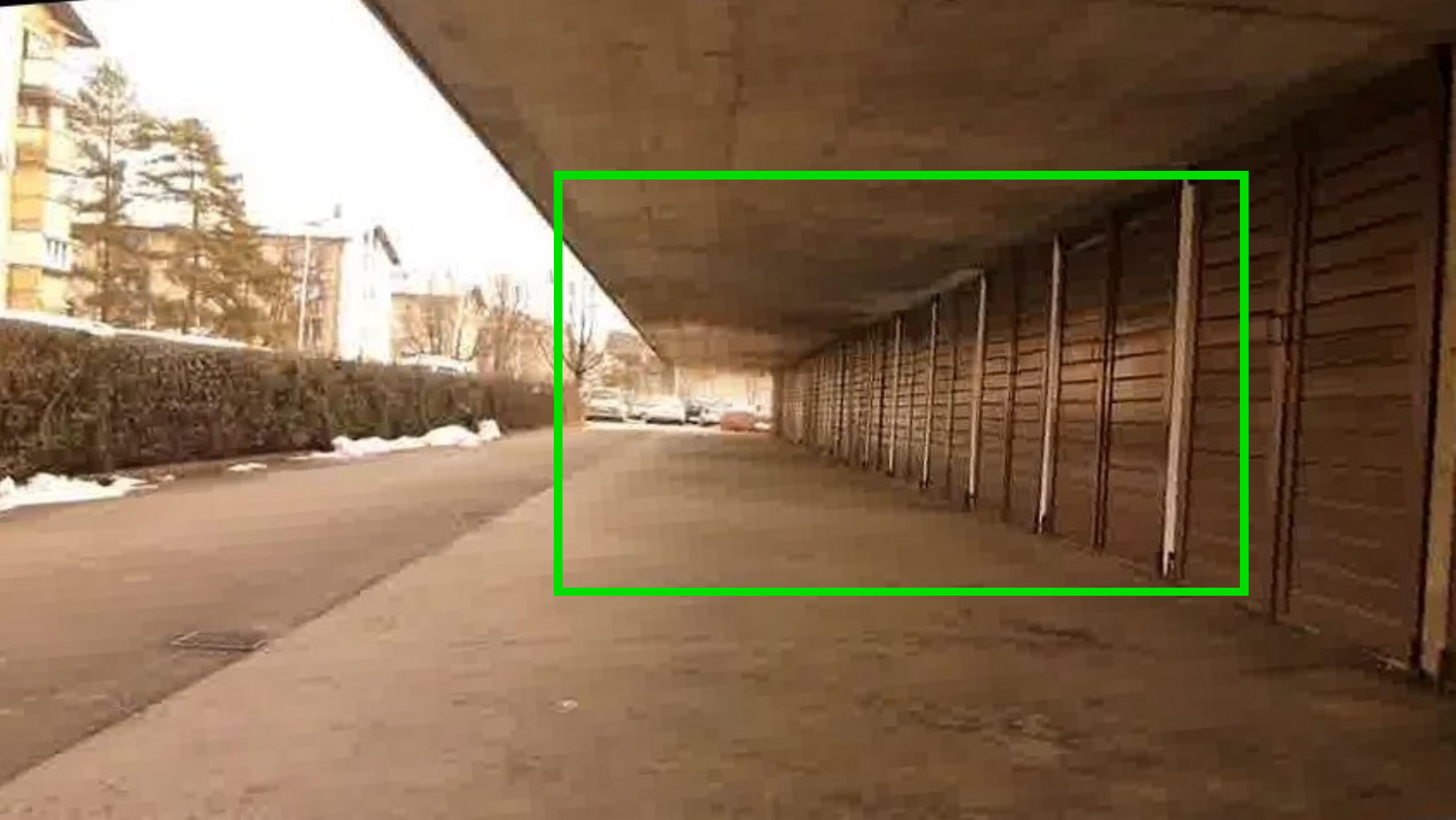}} & 
{\includegraphics[width=0.2\textwidth]{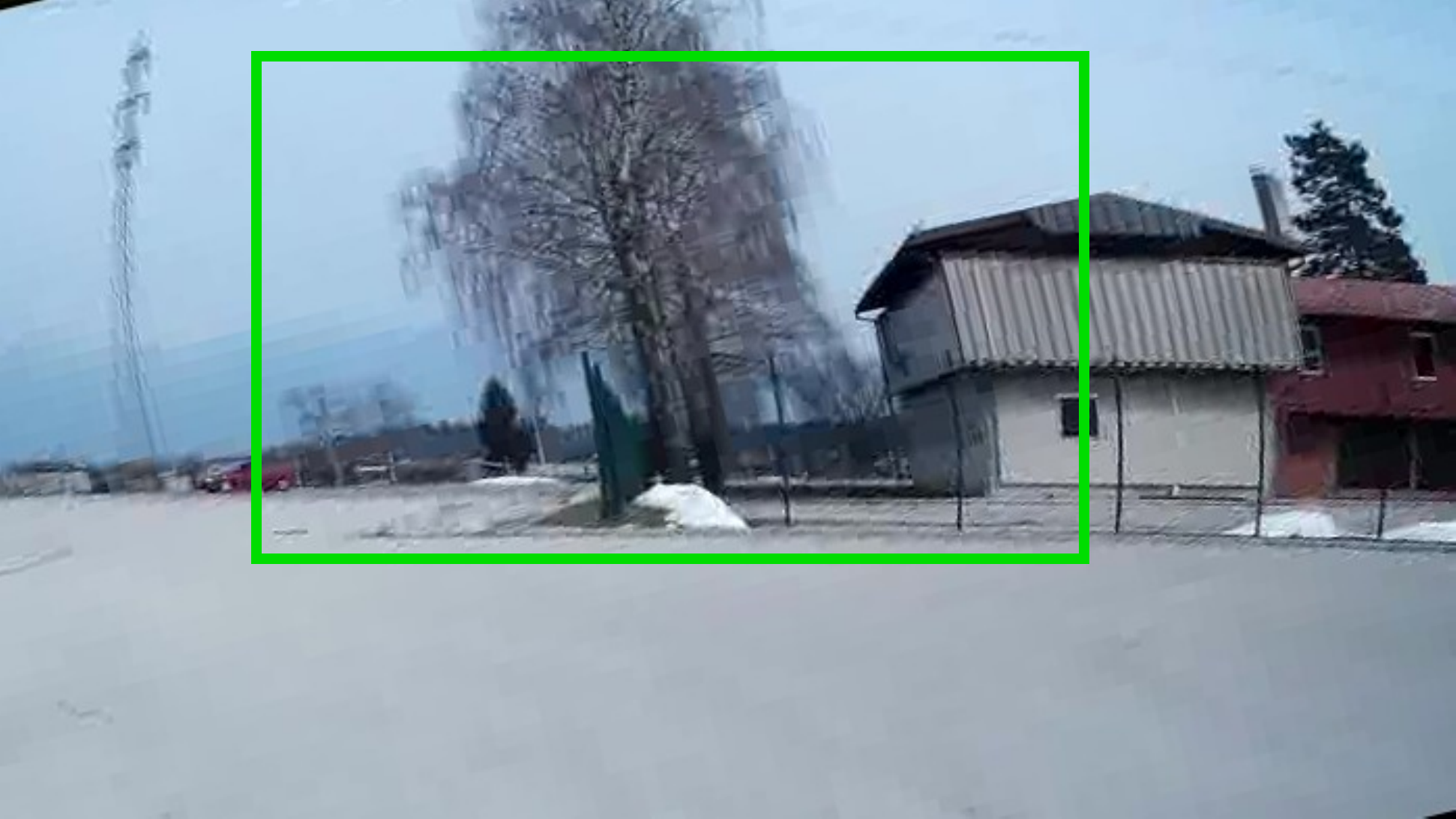}} \\
\begin{picture}(1,25)
  \put(0,0){\rotatebox{90}{~~~~~\color{red}{Output}}}
\end{picture}  & 
{\includegraphics[width=0.2\textwidth]{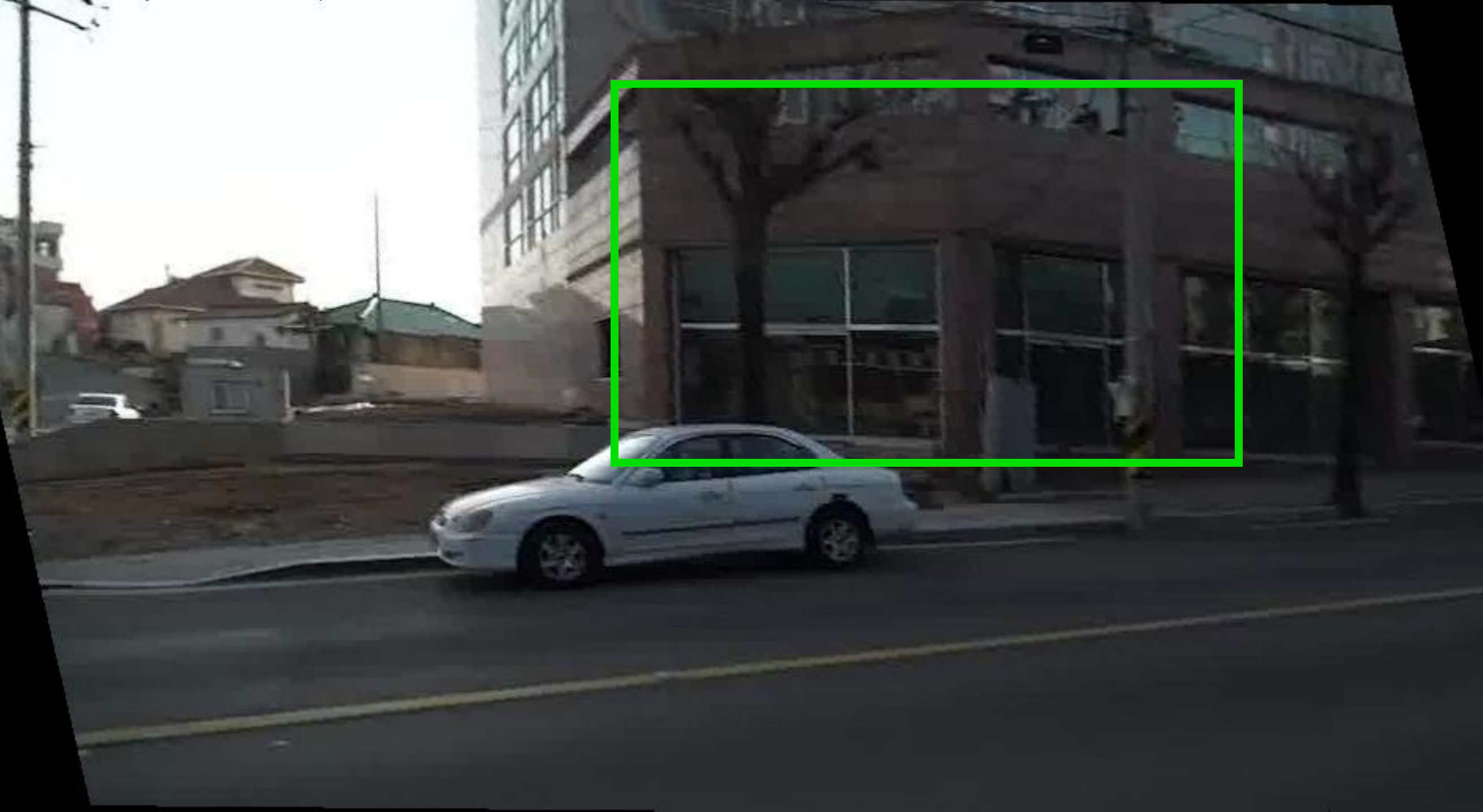}} &
{\includegraphics[width=0.22\textwidth]{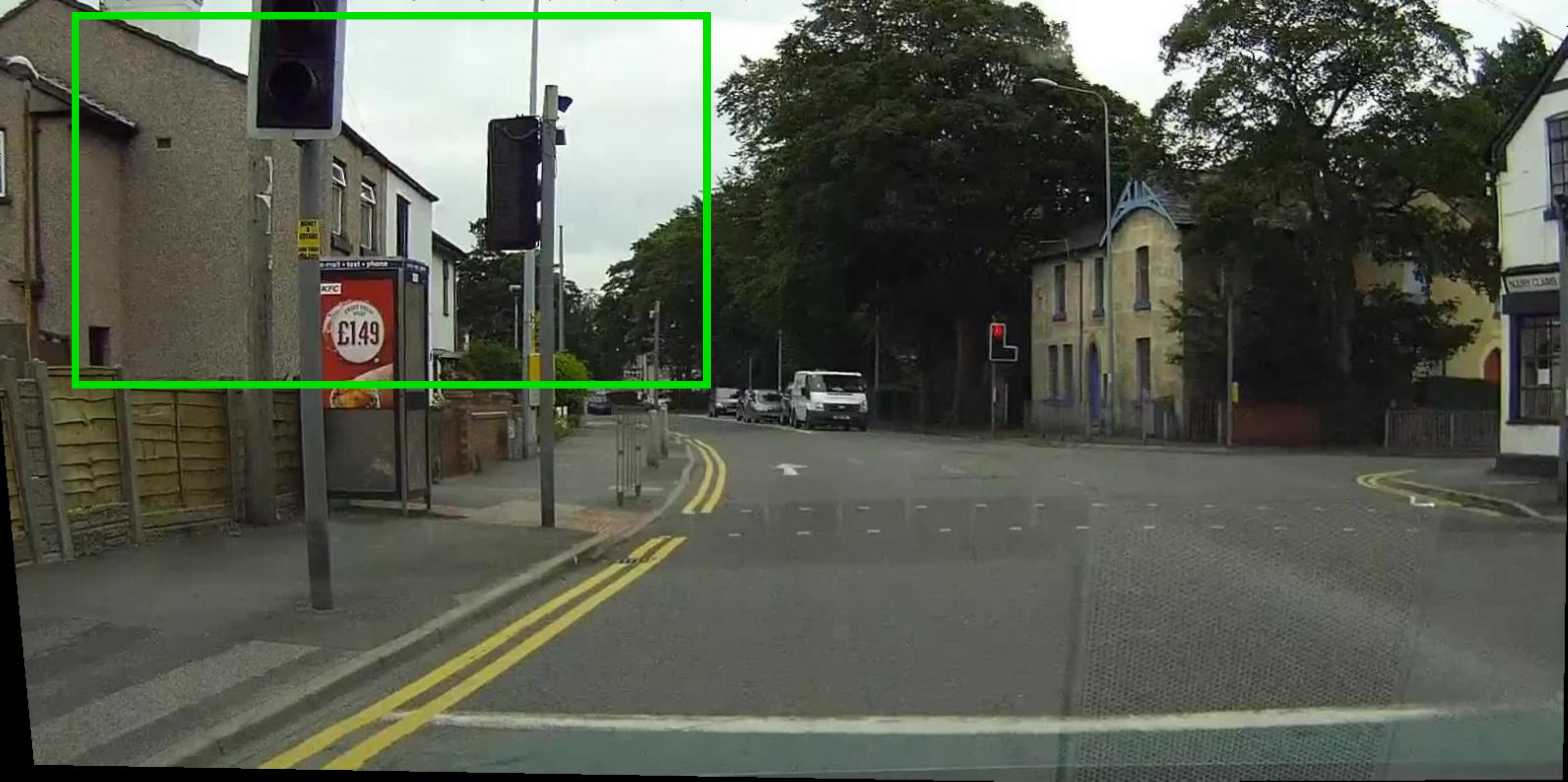}}  & {\includegraphics[width=0.2\textwidth]{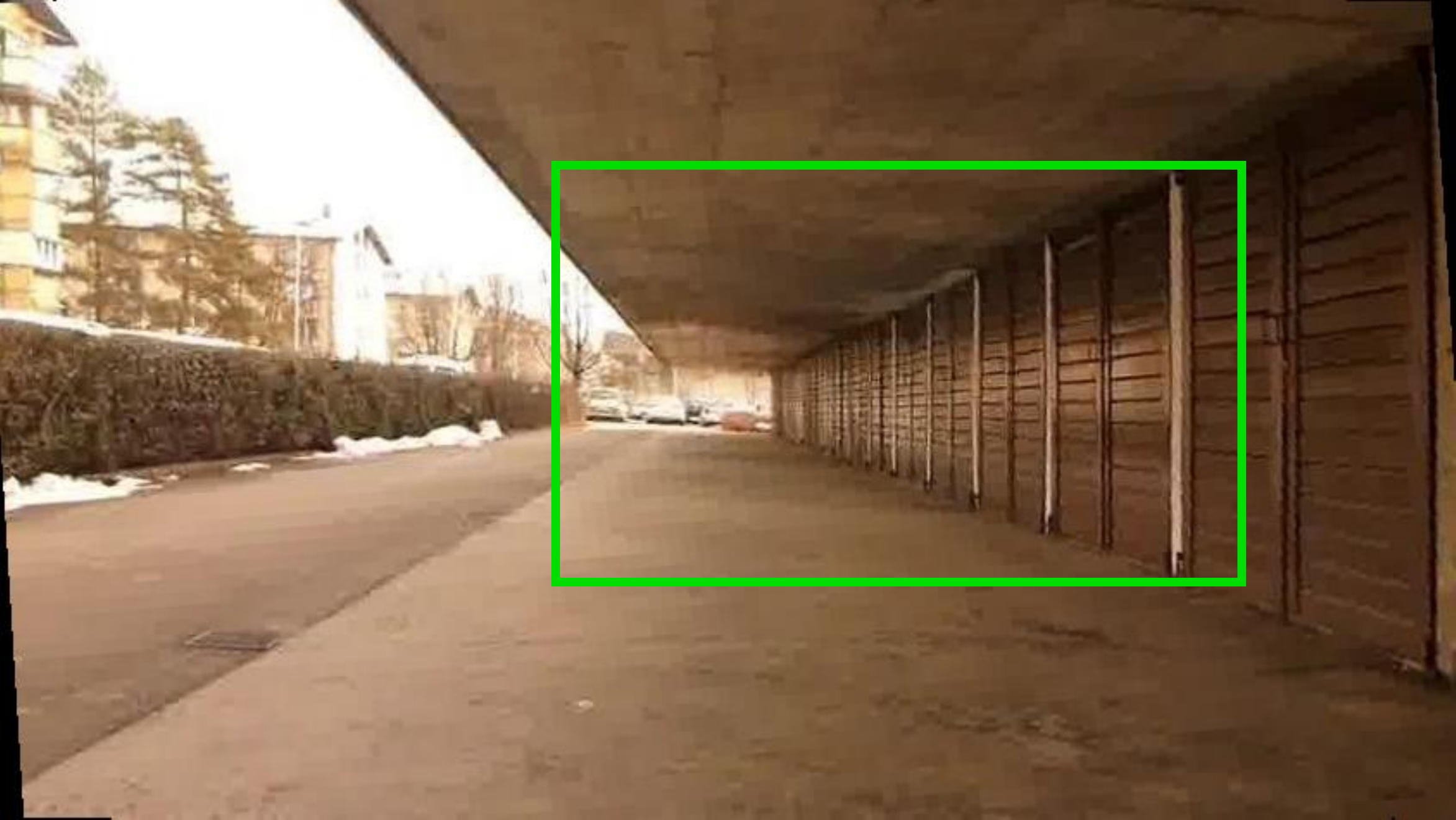}} & 
{\includegraphics[width=0.2\textwidth]{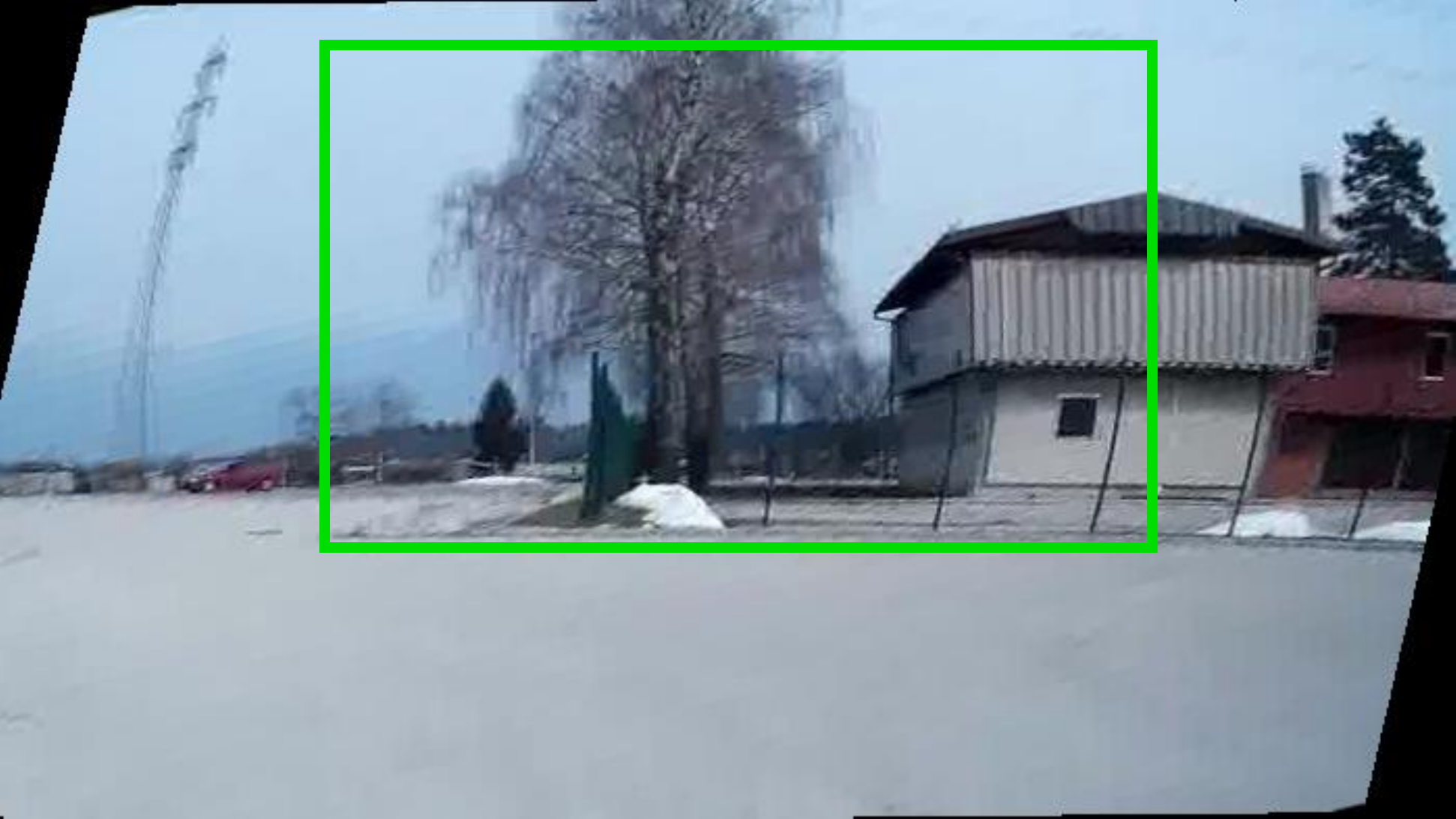}}  \\ 
& (a) SONY-A7S & (b) GS900 & (c) Panasonic-SD9  & (d) Panasonic-SD9 
\end{tabular} 
\begin{tabular}{l@{\hspace{1.3em}}c@{\hspace{0.1em}}c@{\hspace{0.1em}}c@{\hspace{0.1em}}c@{\hspace{0.1em}}c} 
\begin{picture}(1,25)
  \put(0,0){\rotatebox{90}{~~~~~\color{blue}{Input}}}
\end{picture} & 
{\includegraphics[width=0.21\textwidth]{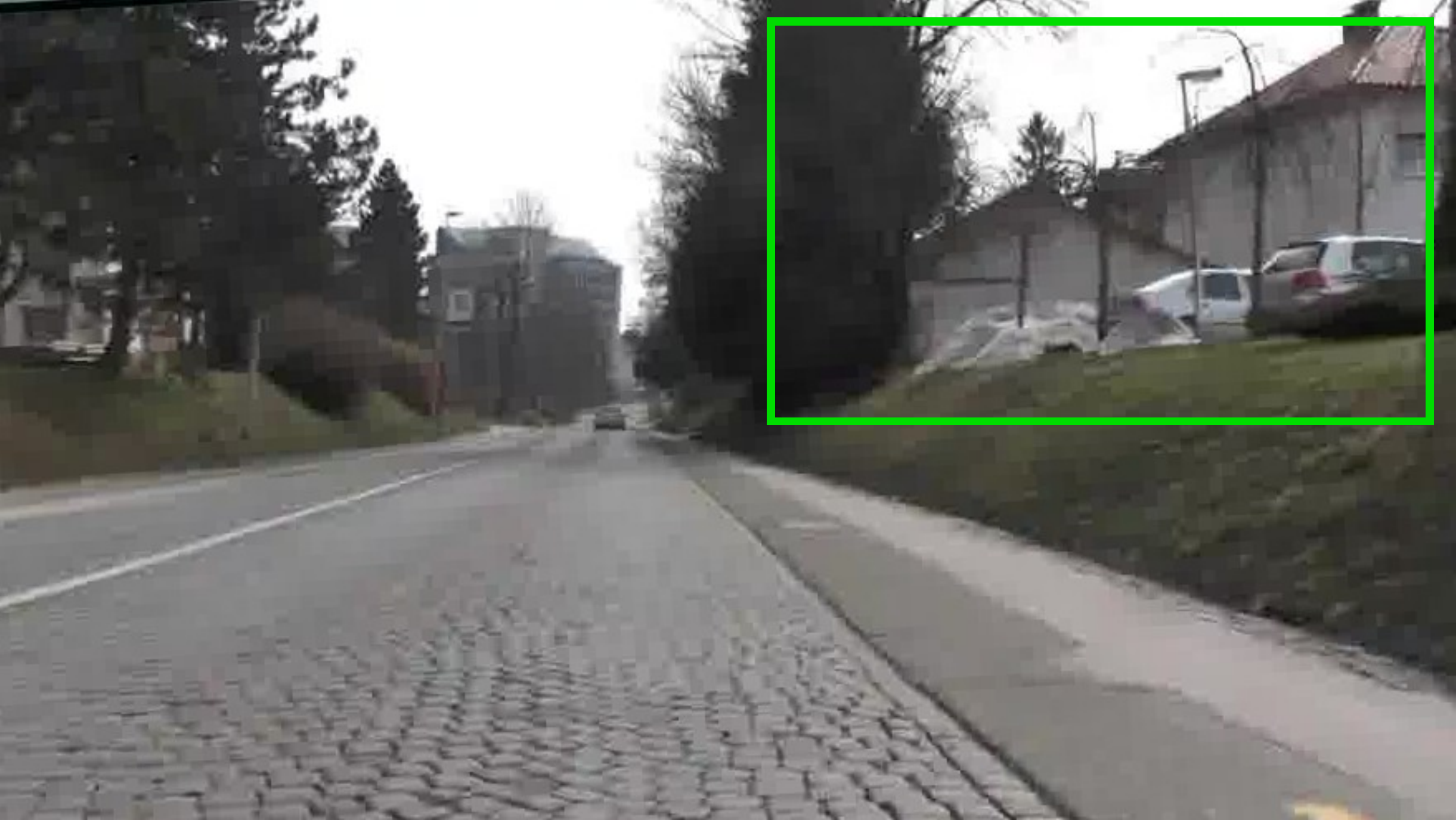}} & 
{\includegraphics[width=0.325\textwidth]{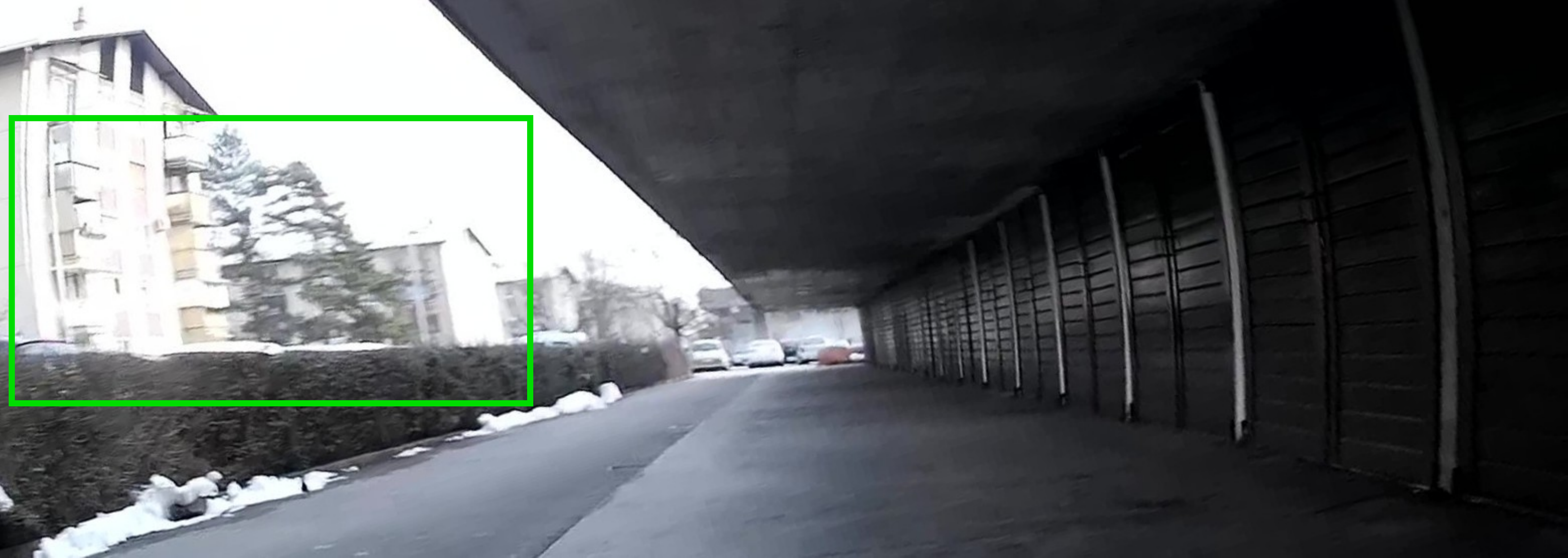}} & 
{\includegraphics[width=0.205\textwidth]{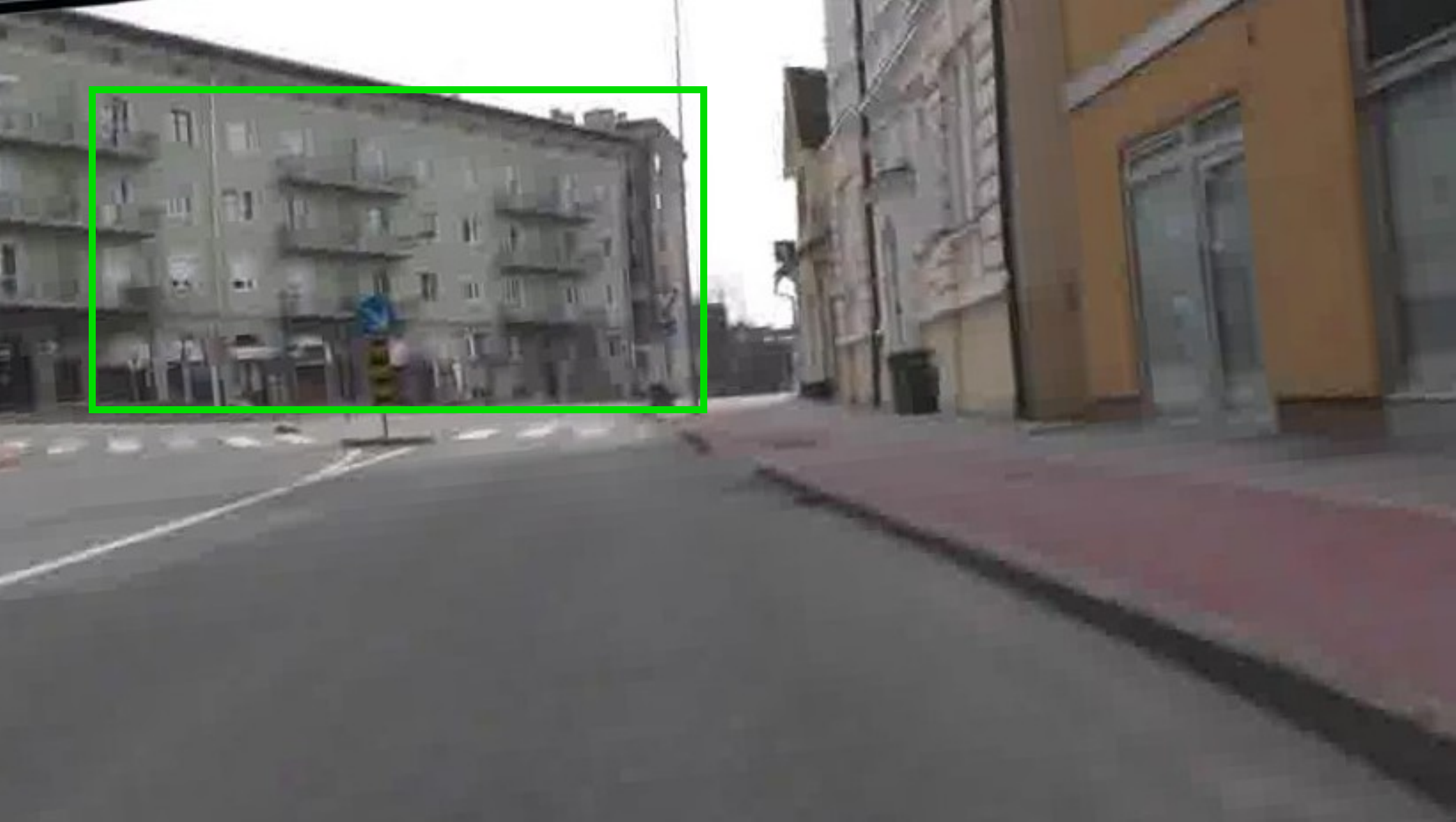}}& 
{\includegraphics[width=0.22\textwidth]{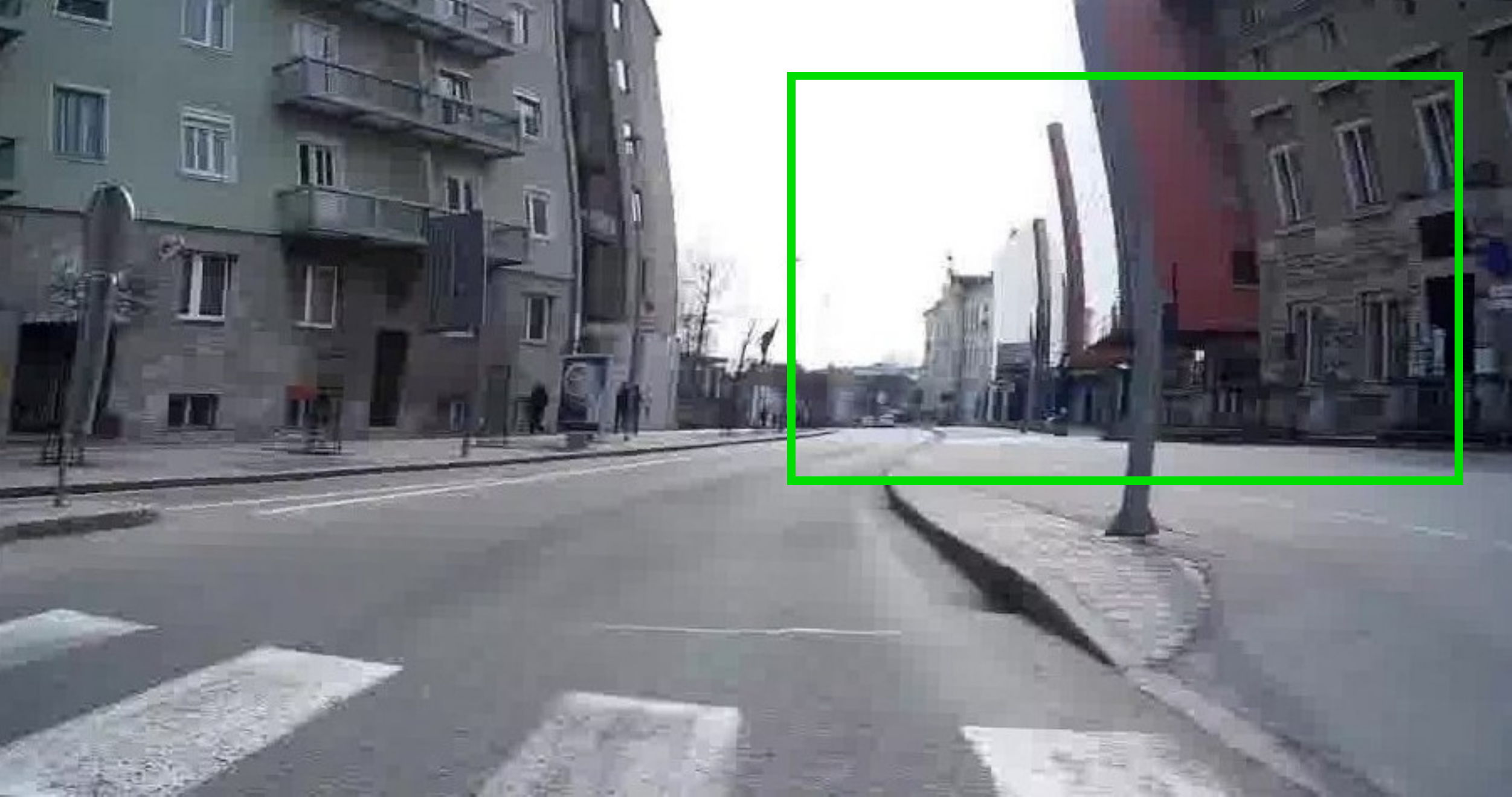}} \\ 
\begin{picture}(1,25)
  \put(0,0){\rotatebox{90}{~~~~~\color{red}{Output}}}
\end{picture} 
& 
{\includegraphics[width=0.21\textwidth]{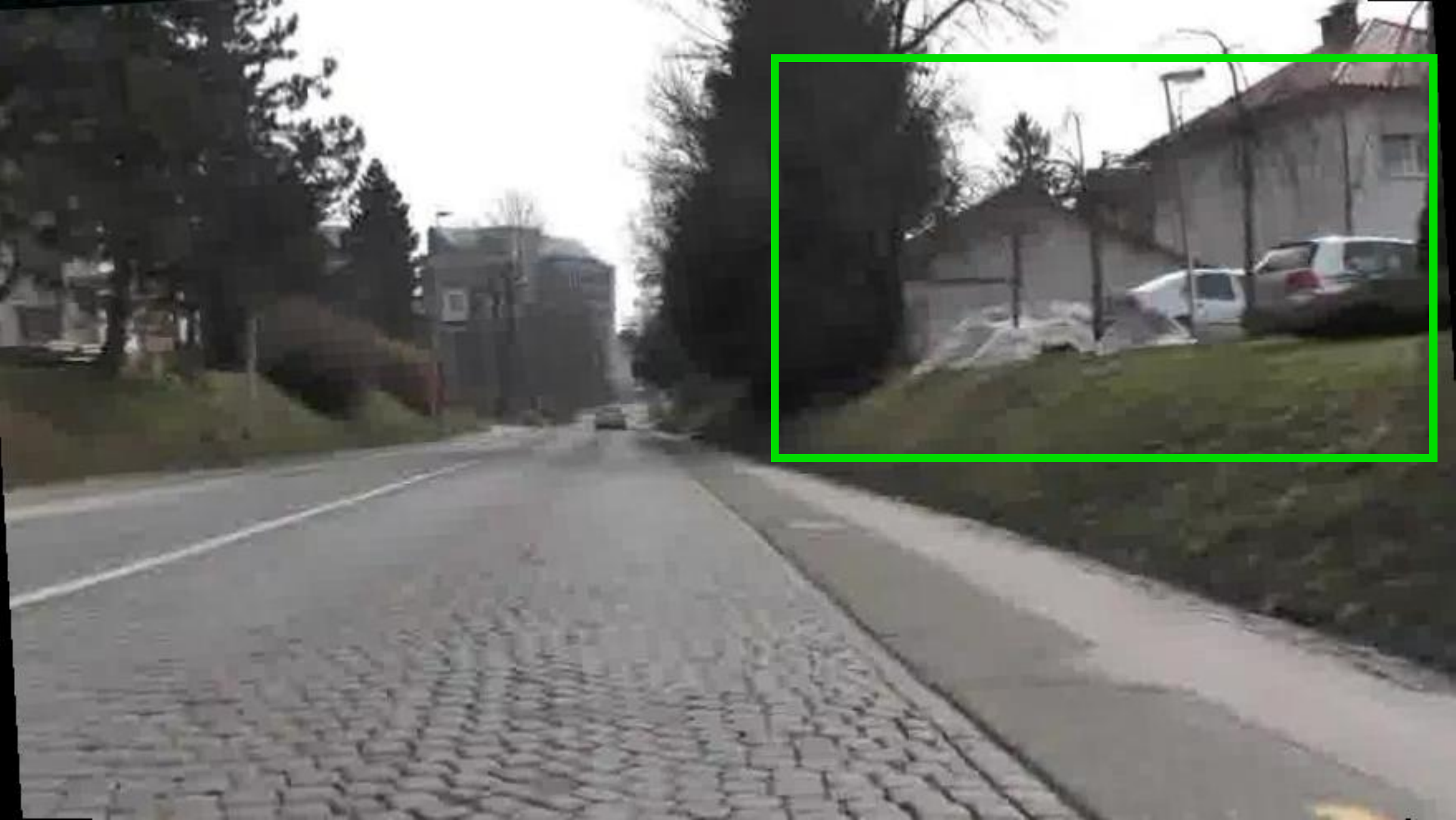}} & 
{\includegraphics[width=0.325\textwidth]{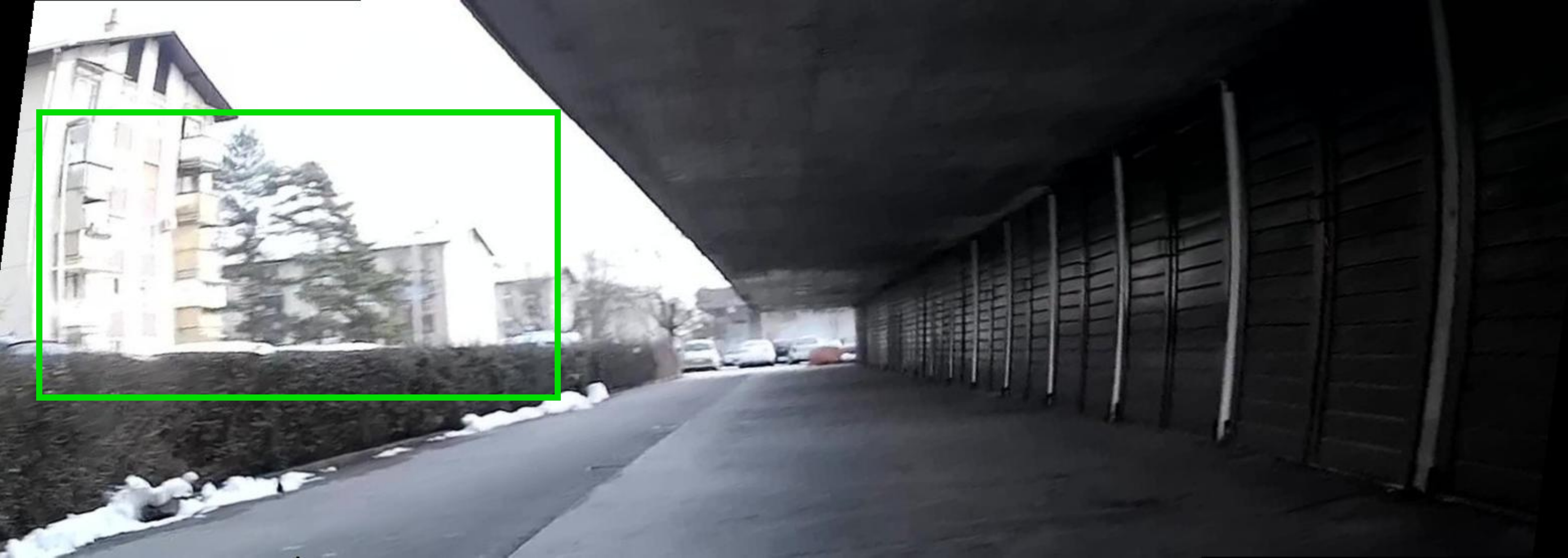}} & 
{\includegraphics[width=0.205\textwidth]{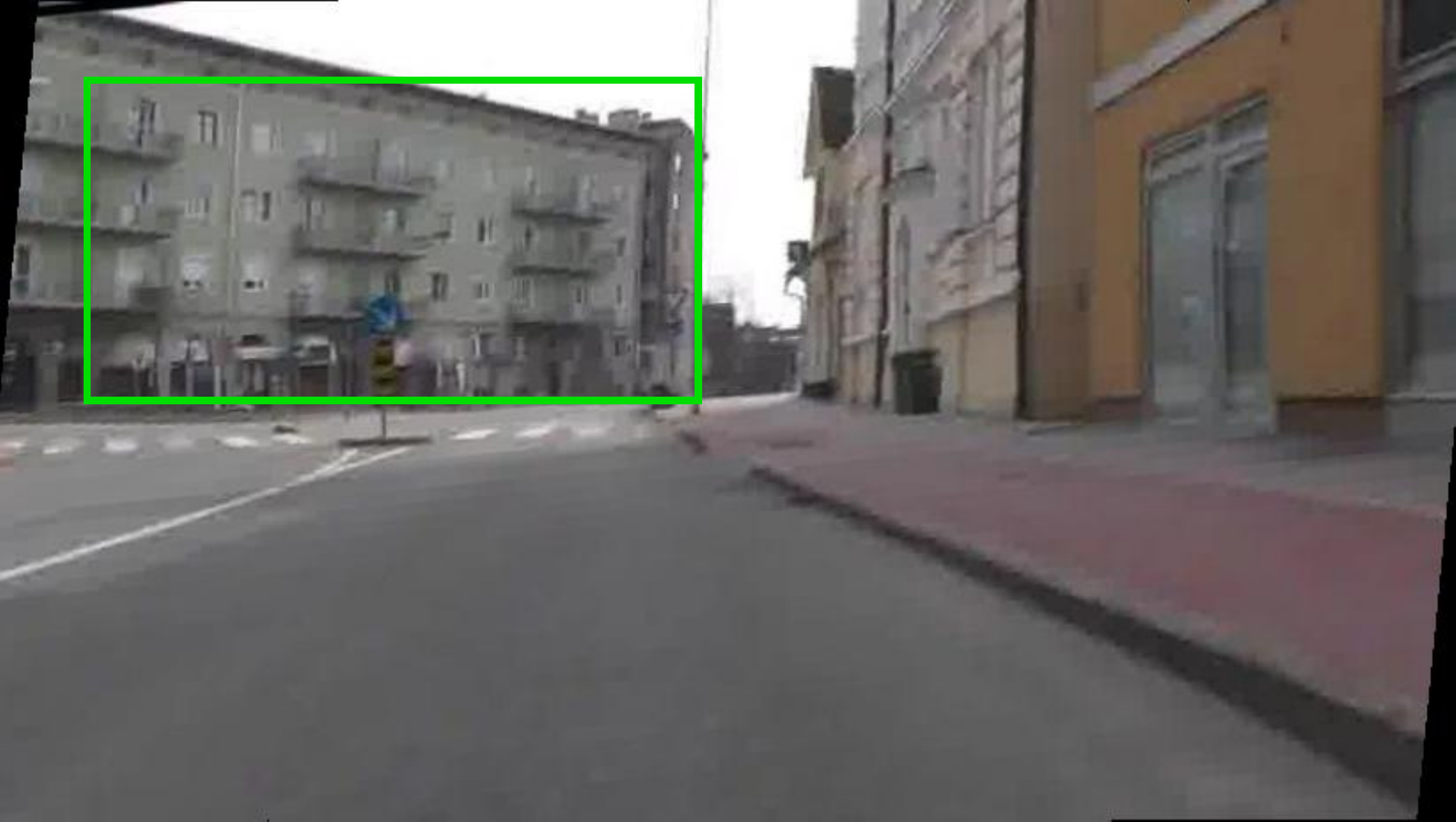}}& 
{\includegraphics[width=0.22\textwidth]{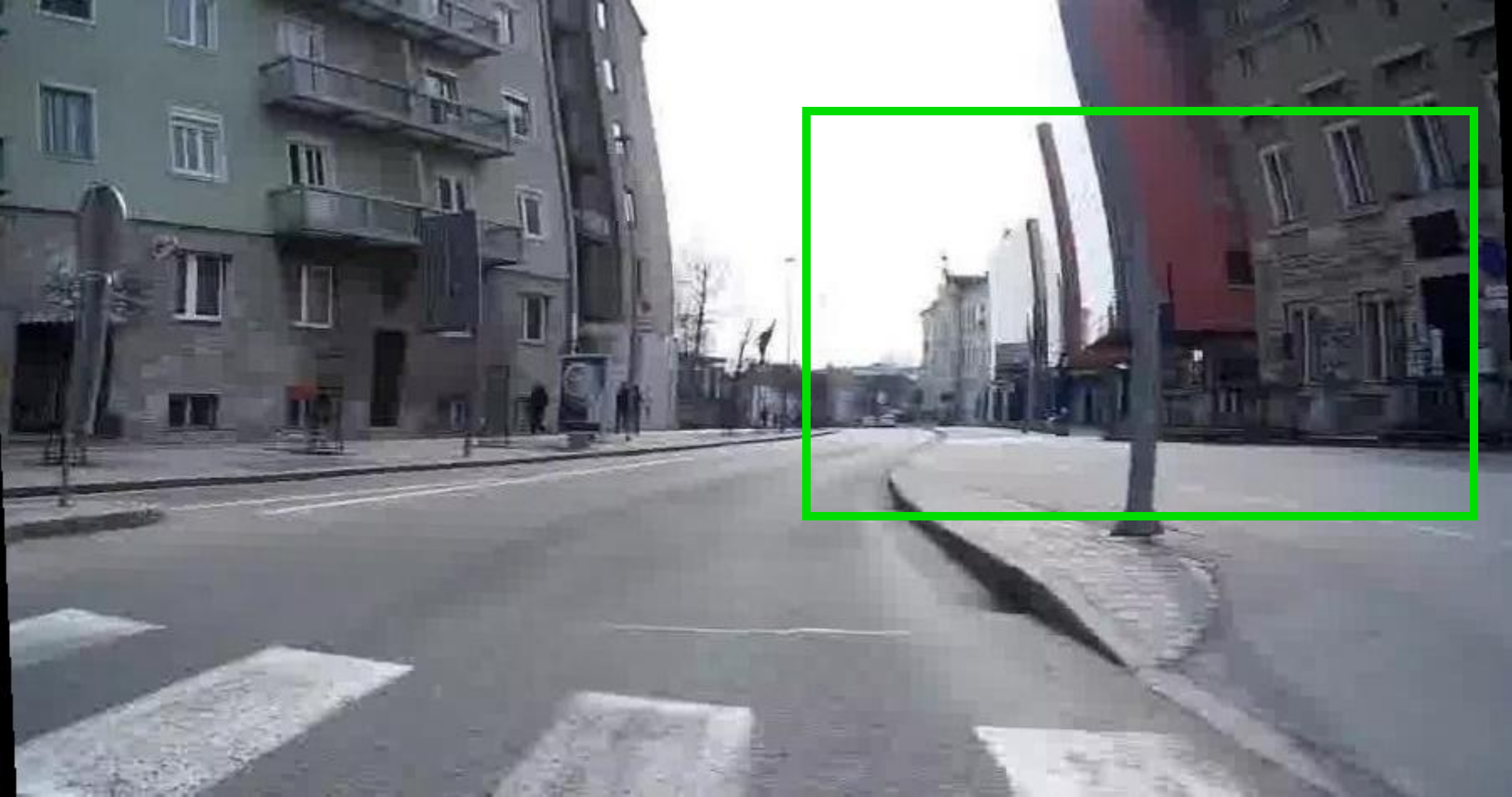}} \\ 
 & (e) Panasonic-SD9 & (f) Panasonic-TS4 & (g) Panasonic-TS4 & (h) Panasonic-SD9 (\color{red}{failure})
\end{tabular}
 \caption{ Evaluation of the proposed method {\bf 4-LA} on real datasets captured by a number of cameras mentioned below. Aesthetically pleasing areas, original and output correspondences, are marked by green rectangular boxes. For visual inspection the readers are encouraged to view the soft copy of the manuscript.} 
  \label{fig:SyntheticMotionEstn26}
\end{figure*}

\section{Experiment}\label{sec:exp}
We conduct experiments on synthetic and real datasets to verify the effectiveness of the proposed minimal solver. Synthetic experiments were aimed to analyze the efficiency and stability of the solver. In particular, the degenerate cases where the camera undergoes a pure translation or rotational velocity. We list our solver as follows 
\begin{itemize}[leftmargin=1em,itemsep=1pt,parsep=1pt]
\item {\bf 4-LA}: The full 4-line solver for Ackermann motion [Gr\"obner method, Eq. \eqref{eq:simplifiedRot8}]. \emph{/* steps}-\ref{4la1}-\ref{4la2}, \emph{algo.}~\ref{algo:RANSAC}\emph{*/}
\item {\bf 3-LA}: The 3-line solver for pure translational motion [Eq.~\eqref{eq:simplifiedRot3}]. \emph{/* steps}-\ref{4la1}-\ref{4la2}, \emph{algo.}~\ref{algo:RANSAC} \emph{modified accordingly */} 
\item {\bf 1-LA}: The 1-line solver for pure rotational motion [Eq.~\eqref{eq:simplifiedRot5}]. \emph{/* steps}-\ref{4la1}-\ref{4la2}, \emph{algo.~}\ref{algo:RANSAC} \emph{modified accordingly */ }
\end{itemize}
\subsection{Parameter settings}\label{sec:paraset}
\noindent We evaluate the proposed solver in conjunction with the {\tt RANSAC} schedule \cite{fischler1981random}. 
In all of our experiments, the inlier threshold is chosen to be $0.5$ pixel and the number of iterations is chosen at a confidence level $0.99$. We utilize the line segment detector \texttt{lsd}~\cite{grompone2010lsd} with all the default parameters. The line segments of size less than $35$ pixels are discarded. 
The line segments are pre-filtered with the algebraic error $|(\bu_i^{rs} \times \bv_i^{rs})^\intercal \be_y | > 0.5$. This increases the efficiency of \texttt{RANSAC} significantly, since it removes clear outliers. 
\subsection{Image Rectification}\label{sec:imgrect} 
\noindent After estimating the parameters, the undistorted image is obtained by the following forward mapping procedure of the RS pixels into the global frame \eqref{eq:RScamera} 
\begin{equation}  
\begin{array}{l}
 \bp~~ = \frac{(1 - \beta p_2^{rs} s\inv)}{(1 - 2p_1^{rs}\alpha t)}(I_3 + 2[\br\scA^t]_\times){\bp}^{rs} + \beta p_2^{rs} s\inv {\bs}\scA^t \\
 s\inv = \max \big( [{p}_1^{rs} - \delta]_- + \lambda [{p}_1^{rs} - \delta]_+ ,\; {p}_2^{rs}\lambda^\ast \big)
\end{array}
\label{eq:simplifiedRot12}
\end{equation} 
where $\br^t\scA = [0, \alpha p_2^{rs}, 0]^\intercal$, ${\bs}\scA^t = [\alpha p_2^{rs}, 0, 1]^\intercal$. Note that $\lambda^\ast$ is the inverse of the height of the camera from the ground plane (assumed to be known). Note that the boundaries of the ground plane and the vertical planes are the pixels for which ${p}_1^{rs} - \delta = {p}_2^{rs}\lambda^\ast, {p}_2^{rs} > 0 $ or $ \lambda ({p}_1^{rs} - \delta) = {p}_2^{rs}\lambda^\ast, {p}_2^{rs} > 0$. 
After the forward map, the unknown pixels are linearly interpolated. Pixels located outside of the compensated frame are placed with intensity $0$. 
 
\subsection{Experiment on synthetic data}\label{sec:syn}
\noindent We choose a clean GS image (size $640 \times 380$, focal length = $816$pixel and principle point at the center of the image) where the camera is placed at $1.2m$ height from the ground plane and $2.5m$ from the vertical plane on the left side of the road. We synthesize RS images with $30$ frames/s ($40\%$ readout time) by simulating the Ackermann motion randomly at a discrete interval of the translational velocity $10-140$km/h and the angular velocity $10-70$deg/s. A random noise is further added with standard deviation of $2$km/h and $2$deg/s while simulating the motion. 

The proposed minimal solvers {\bf 1-LA}, {\bf 3-LA} and  {\bf 4-LA} in conjunction with \texttt{RANSAC} are executed on each of the instances of the synthetic RS image. The statistics of the estimated motions are plotted in Figure~\ref{fig:SyntheticMotionEstn2}. We observe that overall the motions are estimated accurately.  Moreover, the joint estimations of the velocities are as accurate as individual estimations. Note that for a small velocity, the estimation of the line at infinity (and therefore the scene depth) is not accurate, thus, the estimation is not very accurate for small velocity. 
However, in almost all the cases, the proposed minimal solvers produce visually plausible images. In Figure~\ref{fig:synthetic_data}, we display few results of such instances. We observe that in some cases our method does not estimate the rolling shutter motion accurately nonetheless it improves the image.   

\subsection{Robustness against our assumptions}\label{sec:syn2}
\noindent {\bf Known verticle angle:} In Figure~\ref{fig:SyntheticMotionEstn2}(g), we run experiments to show the robustness of our method under the known vertical angle. In this experiment, we have generated $100$ random vertical direction with
normal distribution for maximal deviation of $1$deg to
simulate industry quality IMU reading error \cite{kukelova2010closed} and observe minimal effect on the results. \\ 
\noindent {\bf Known scale:} In Figure \ref{fig:SyntheticMotionEstn2}(h), we show the motion prediction under the scale (distance of one of the vertical plane from the camera) error. However, as the scale is used only to compute the car motion in units and there will not be any effect on the output image quality. \\ 
\noindent {\bf Known height of the camera:} Moreover, we also observe minimal effect on the output image quality (as shown in Fig.~\ref{fig:synthetic_data22}), if the height of the camera from the ground plane is not given accurately. 

 
\subsection{Experiment on real data} 
\noindent We employ the proposed method {\bf 4-LA} on a number of rolling shutter distorted image sequences, downloaded from YouTube\footnote{\url{https://www.youtube.com/}}, and the results are displayed in Figure \ref{fig:SyntheticMotionEstn26}. The only qualitative evaluation is conducted as no ground-truths are available. 
The dataset is uncalibrated, thus, we estimated the focal length as $0.9$ times the maximal image dimension. The principal point was  chosen to be the center of the image. Overall, proposed method performs quite well for all the images. However, the car was shaking very rapidly for Figure \ref{fig:SyntheticMotionEstn26}(h) and proposed method could not perform very well as in this scenario the car motion do not quite follow the assumed Ackemann motion.    

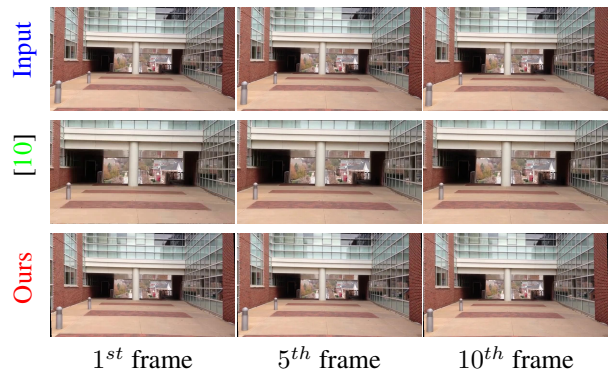
\begin{figure}
\centering 
\begin{tabular}{l@{\hspace{1.3em}}c@{\hspace{0.1em}}c@{\hspace{0.1em}}c@{\hspace{0.1em}}c@{\hspace{0.1em}}c} 
\begin{picture}(1,25)
  \put(0,0){\rotatebox{90}{~~~~~\color{blue}{Input}}}
\end{picture} & 
{\includegraphics[width=0.14\textwidth]{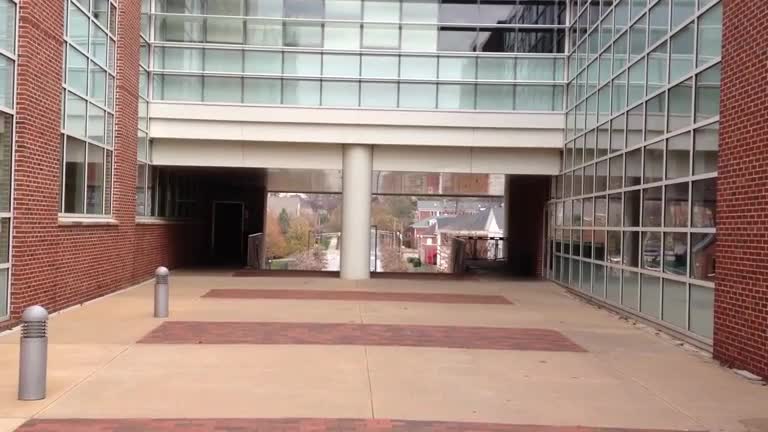}} &
{\includegraphics[width=0.14\textwidth]{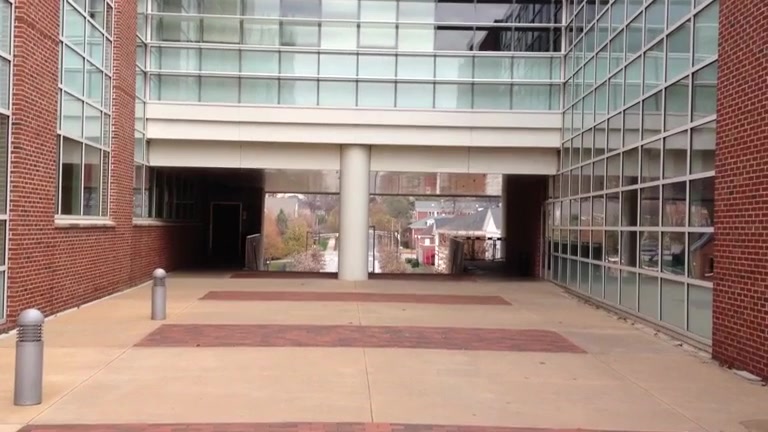}} & 
{\includegraphics[width=0.14\textwidth]{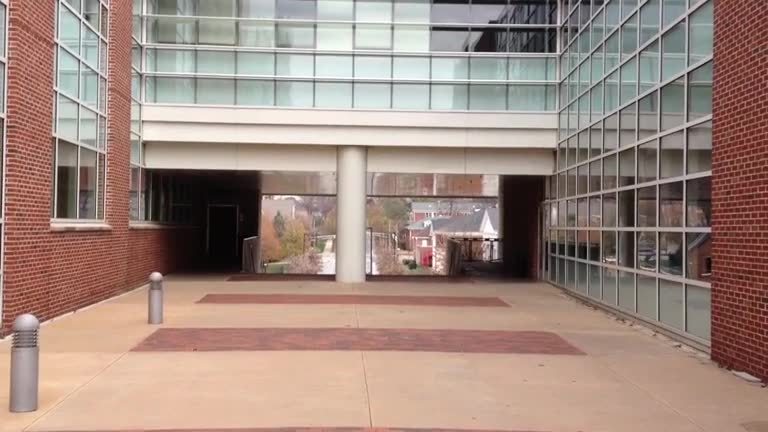}} \\
\begin{picture}(1,25)
  \put(0,0){\rotatebox{90}{~~~~~~~\cite{grundmann2012calibration}}}
\end{picture} & 
{\includegraphics[width=0.14\textwidth]{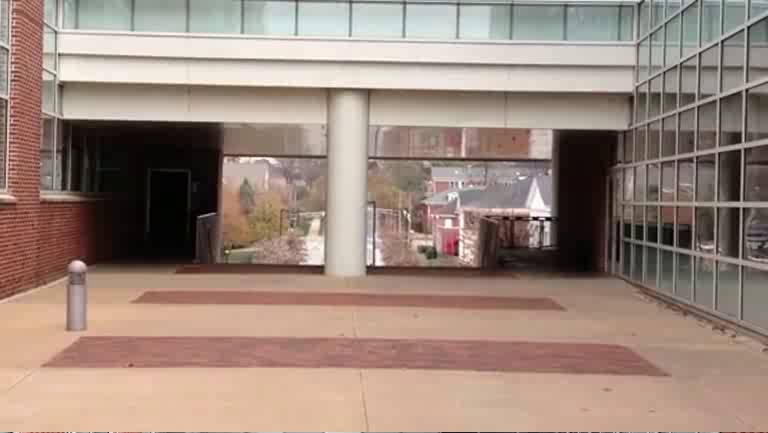}} &
{\includegraphics[width=0.14\textwidth]{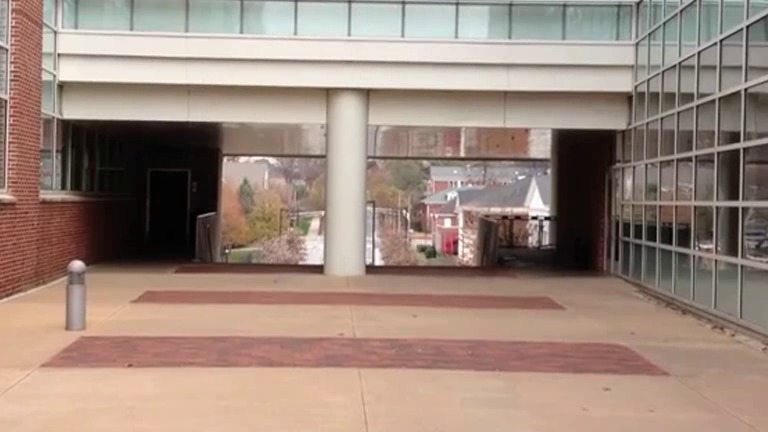}} & 
{\includegraphics[width=0.14\textwidth]{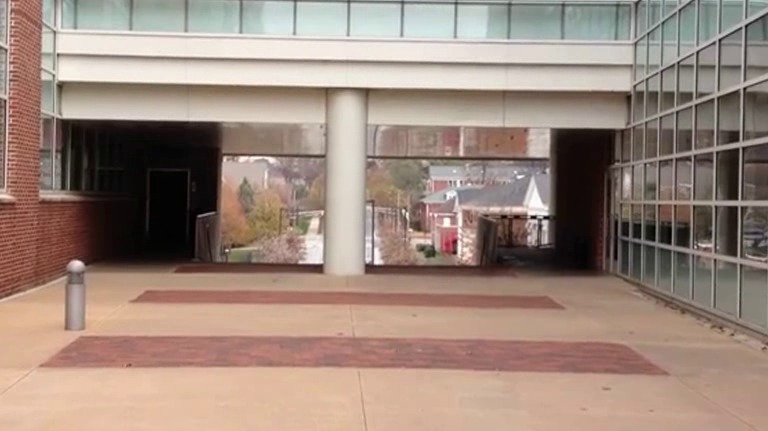}} \\
\begin{picture}(1,25)
  \put(0,0){\rotatebox{90}{~~~~~\color{red}{Ours}}}
\end{picture}  & 
{\includegraphics[width=0.14\textwidth]{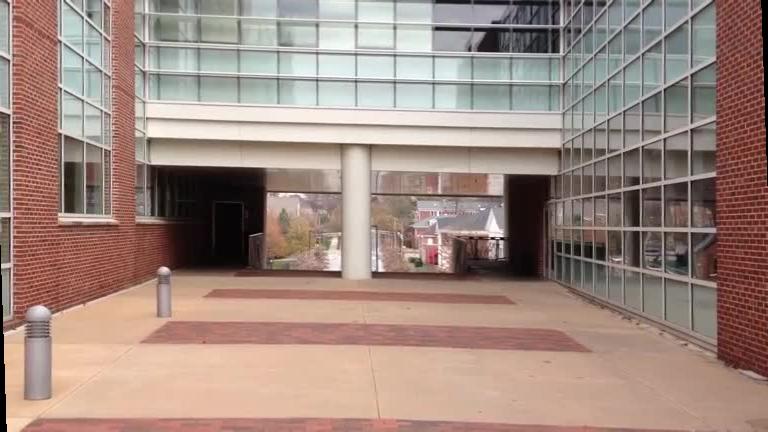}} &
{\includegraphics[width=0.14\textwidth]{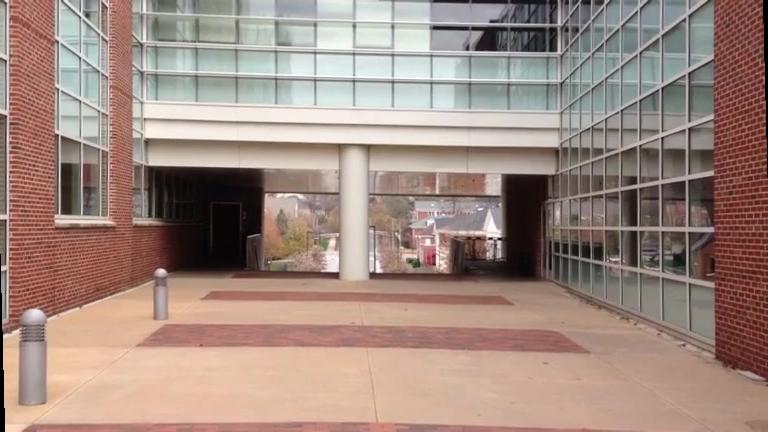}} & 
{\includegraphics[width=0.14\textwidth]{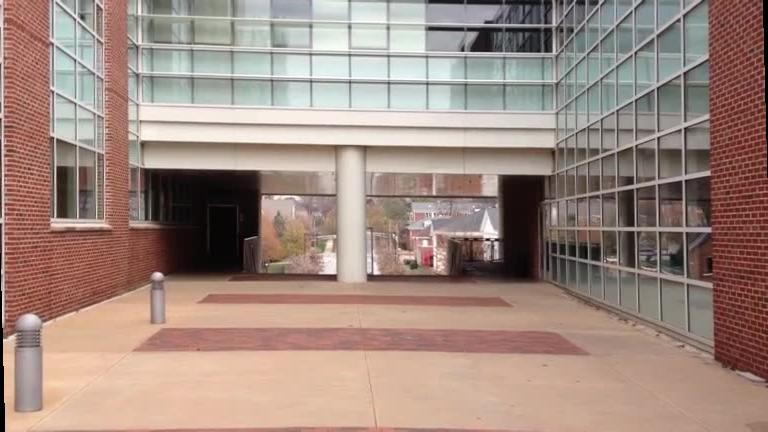}} \\
& $1^{st}$ frame & $5^{th}$ frame & $10^{th}$ frame
\end{tabular}
 \caption{ Evaluation of the proposed method {\bf 4-LA} on a real data sequence \texttt{i4s$\_$depth$\_$3$\_$.mp4} from the dataset \cite{grundmann2012calibration} captured by iPhone4. We display 1st, 5th and 10th images of the sequence. } 
  \label{fig:SyntheticMotionEstn24}
\end{figure}

The proposed method was also evaluated frame-by-frame on the image sequences from the datasets~\cite{grundmann2012calibration}\footnote{\url{http://www.cc.gatech.edu/cpl/projects/rollingshutter/}}.  The results are displayed in Figure \ref{fig:SyntheticMotionEstn24}. Note that  \cite{grundmann2012calibration} do not consider the translational motion while compensating the motion and the resulted image were cropped for visually pleasant output. This dataset was chosen for evaluation as it consists of corridor scene required by the proposed algorithm, despite captured by an handhold camera and there is little rolling shutter effect.  Readers are requested to look at the zoomed version of the draft. 

\begin{figure}
\centering  
\begin{tabular}{c@{\hspace{0.1em}}c@{\hspace{0.1em}}c} 
{\includegraphics[width=0.15\textwidth]{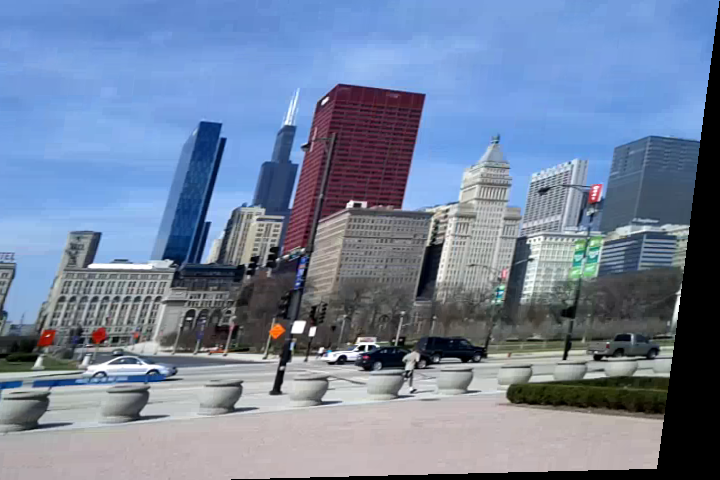}} & 
{\includegraphics[width=0.15\textwidth]{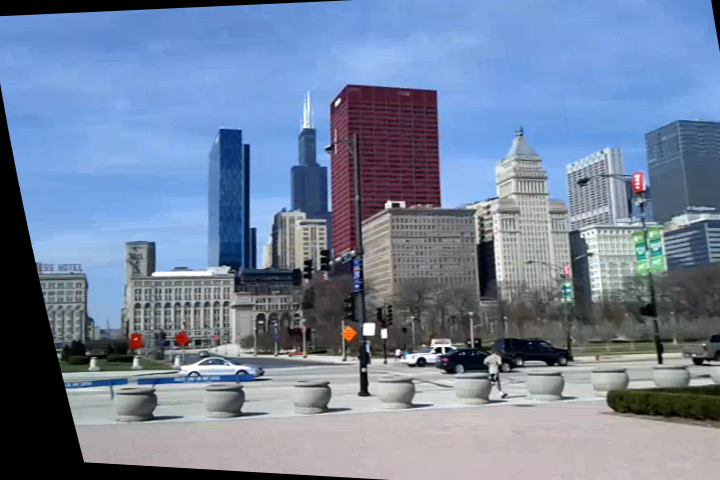}} & 
{\includegraphics[width=0.15\textwidth]{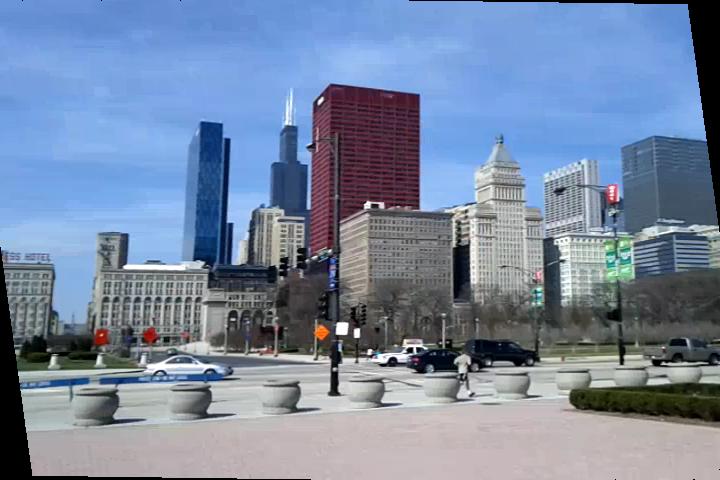}} \\ 
(a) \cite{jia2012probabilistic} & (b) \cite{rengarajan2016bows}  & (c) \color{red}{Ours}  
\end{tabular} 
 \caption{Comparison of proposed {\bf 1-LA} on a real video sequence. Note that proposed method estimates only $yaw$ rotation in contrast \cite{rengarajan2016bows} estimates full rotation with the cost of high computation time mentioned in Table \ref{tab:runtime}} 
  \label{fig:RealMotionEstn2} 
\end{figure}

In a different settings, 
we choose the real dataset \cite{gyrodataset} where the dataset was captured from a fixed location only moving the camera horizontally. Thus this is the scenario of a specific motion where translation is negligible. Here, we evaluate our algorithm {\bf 1-LA} with baselines \cite{jia2012probabilistic,rengarajan2016bows} and the results are displayed in Figure~\ref{fig:RealMotionEstn2}. 
Note that \cite{jia2012probabilistic} compensates the camera motion by the gyroscope readings. \cite{rengarajan2016bows} estimates the yaw and roll angular velocities where as our solver estimates only the yaw angular velocity. However, {\bf 1-LA} produces equivalent  qualitative results on the selected images while three orders of magnitude faster than the baseline \cite{rengarajan2016bows}.

\subsection{Runtime comparison} 
\noindent The proposed method is implemented in \texttt{Matlab} and the runtimes are computed on an $i7$ $2.8$GHz CPU utilizing only a single core.  On average it takes around $0.087$ second to rectify a $360 \times 520$ image, excluding LS detection and the rectification (also exclude from \cite{grundmann2012calibration}, \cite{rengarajan2016bows} and \cite{Su_2015_CVPR}), which is two-three orders of magnitude faster over the most recent methods. See Table \ref{tab:runtime} for the detailed comparison. Note that we rescale the runtime comparison according to the image size reported in the paper. 
\begin{table}[!ht]
\caption{Average runtime of different methods in seconds.}
\centering
\begin{tabular}{c@{\hskip 0.14in}c@{\hskip 0.14in}c@{\hskip 0.14in}c@{\hskip 0.14in}c@{\hskip 0.14in}c@{\hskip 0.14in}c} \hline \hline 
Methods & {\bf 1-LA} & {\bf 3-LA} & {\bf 4-LA} & \cite{grundmann2012calibration} & \cite{rengarajan2016bows} & \cite{Su_2015_CVPR}  \\  
Runtime & 0.001s & 0.003s & 0.087 & 0.1s & 8s & 249.6s \\ \hline 
\end{tabular}
\label{tab:runtime}
\end{table}
\section{Conclusion }\label{sec:conclude}
We proposed a minimal solver for rolling shutter camera motion compensation under an Ackermann motion from a single view. A 4-line algorithm is developed to estimate the simplified depth of the scene along with the motion parameters. The proposed method is also the first of its kind to exploit the minimal solvers for monocular rolling shutter motion compensation which allows to have an extremely fast compensation method. Extensive experiments were performed on the simulated and real datasets. The results demonstrate the computational efficiency and the effectiveness of the proposed approach. 
During the motion compensation in a video, individual frames were compensated separately; however, a nonlinear optimization method (viz., rolling shutter bundle adjustment) can be developed for more accurate estimation with proposed fast method as an initialization and hence there lies a potential future extension. 
\subsubsection*{Acknowledgements} We thank V. Rengarajan for running his method \cite{rengarajan2016bows} and providing the results on selected datasets [Fig. \ref{fig:RealMotionEstn2}].

{\small
\bibliographystyle{ieee}
\bibliography{egbib}
}

\end{document}